\documentclass{ieeeaccess}
\usepackage{cite}
\usepackage{amsmath,amssymb,amsfonts}
\usepackage{algorithmic}
\usepackage{graphicx}
\usepackage{textcomp}
\usepackage{float}
\usepackage{url}
\usepackage{bm}
\usepackage{booktabs} 
\usepackage{multirow} 
\newcommand{\ra}[1]{\renewcommand{\arraystretch}{#1}}
\makeatletter
\AtBeginDocument{\DeclareMathVersion{bold}
\SetSymbolFont{operators}{bold}{T1}{times}{b}{n}
\SetSymbolFont{NewLetters}{bold}{T1}{times}{b}{it}
\SetMathAlphabet{\mathrm}{bold}{T1}{times}{b}{n}
\SetMathAlphabet{\mathit}{bold}{T1}{times}{b}{it}
\SetMathAlphabet{\mathbf}{bold}{T1}{times}{b}{n}
\SetMathAlphabet{\mathtt}{bold}{OT1}{pcr}{b}{n}
\SetSymbolFont{symbols}{bold}{OMS}{cmsy}{b}{n}
\renewcommand\boldmath{\@nomath\boldmath\mathversion{bold}}}
\makeatother

\def\BibTeX{{\rm B\kern-.05em{\sc i\kern-.025em b}\kern-.08em
    T\kern-.1667em\lower.7ex\hbox{E}\kern-.125emX}}

\begin{document}
\history{This work has been submitted to the IEEE for possible publication. Copyright may be transferred without notice, after which this version may no longer be accessible}
\doi{}

\title{BIOLOGICAL SEX DETERMINATION IN CADAVERS USING DEEP LEARNING ALGORITHMS FROM COMPUTED TOMOGRAPHY IMAGES OF PELVIS AND SKULL}
\author{\uppercase{Giovanna Herculano Tormena}\authorrefmark{1}, 
\uppercase{Davi Nascimento Araújo}\authorrefmark{1}, 
\uppercase{Germano Coimbra Soares de Carvalho}\authorrefmark{1},
\uppercase{Gustavo Bruno Centenaro}\authorrefmark{1}, 
\uppercase{Rafael Janowski Pozzer}\authorrefmark{1}, 
\uppercase{Rodrigo Akira Azevedo Kurosawa}\authorrefmark{1}, 
\uppercase{Danilo Aires Alves}\authorrefmark{2}, 
\uppercase{Filipe Thiago Xavier de Campos}\authorrefmark{2}, 
\uppercase{Pedro Henrique Macedo dos Santos}\authorrefmark{2}, 
\uppercase{Pedro Augusto Prado Mota}\authorrefmark{2}, 
\uppercase{Ricardo V. Godoy}\authorrefmark{1}, \IEEEmembership{Member,~IEEE},
\uppercase{João Manoel Herrera Pinheiro}\authorrefmark{1}, \IEEEmembership{Graduate Student Member,~IEEE}, and
\uppercase{Marcelo Becker}\authorrefmark{1}, \IEEEmembership{Member,~IEEE}}

\address[1]{São Carlos School of Engineering, University of São Paulo, São Carlos 13566-590, Brazil}
\address[2]{Aristoclides Teixeira Institute of Forensic (IMLAT), Goiânia 74.425-030, Brazil}
\tfootnote{This work was supported by the Foundation for the Support of Physics and Chemistry - FAFQ. The Article Processing Charge (APC) for the publication of this research was funded by the Coordination for the Improvement of Higher Education Personnel - CAPES (ROR identifier: 00x0ma614). For the purpose of open access, the authors have applied a Creative Commons CC BY license to any Author Accepted Manuscript version arising from this submission. }

\markboth
{G. H. Tormena \headeretal: Biological sex determination in cadavers using deep learning algorithms}
{G. H. Tormena \headeretal: Biological sex determination in cadavers using deep learning algorithms}

\corresp{Corresponding author: Giovanna Herculano Tormena (e-mail: giovannatormena@usp.br).}

\begin{abstract}
Sexual identification of decomposed cadavers challenges traditional methods dependent on visual anthropological analysis. This study evaluates state-of-the-art deep learning (including YOLO26, YOLO11, ConvNeXt-Tiny, EfficientNetV2, ViT-B16, VGG16, and ResNet50) with transfer learning to automatically determine biological sex from forensic computed tomography (CT) scans. We analyzed 141 autopsied cadavers from the Forensic Medical Institute of Goiânia-GO, including a broad age range and varying conditions of preservation. The three-dimensional reconstructions of the pelvis and skull were converted into standardized two-dimensional profile projections, contributing to the study of this new technical approach. Data augmentation techniques compensated for sample limitations. Two scenarios were validated: binary and quaternary classification (one class per sex vs. one class per anatomical region of each sex). The best-performing model achieved highly consistent results on the pelvis region and still satisfactory performance on the skull region, reaching an overall patient-level accuracy of 95.65\%, recall of 92.86\%, F1-score of 94.36\%, and precision of 97.22\%, maintaining consistent performance across the evaluated cases, including those with trauma-related artifacts. Results indicate the technical feasibility of the methodology, demonstrating that deep learning models can provide objective, high-speed skeletal analysis. Since the study was conducted using data from a single institution and a single computed tomography scanner, further validation across multiple centers and scanners is required to assess the generalizability of the proposed approach.

\end{abstract}


\titlepgskip=-21pt

\maketitle

\section{Introduction}
\label{sec:introduction}
\PARstart In the field of Forensic Anthropology, the determination of biological sex is a fundamental step in building an individual's biological profile and remains a cornerstone of human identification. Traditionally, this process relies on a morphological and metric analysis of skeletal structures with pronounced sexual dimorphism, particularly the skull and pelvis \cite{20deantes}. While effective, classical methods present significant limitations: they are observer-dependent, time-consuming, and their performance decreases in challenging scenarios, such as mass disasters or in cases involving charred, fragmented, or severely decomposed remains~\cite{krishan2016review, seifert2017applicability, 5deantes}. Given these challenges, Artificial Intelligence (AI) has emerged as a high-potential tool to transform forensic analysis~\cite{1deantes, galante2023}.

In the specific context of medical imaging, deep learning architectures are widely employed~\cite{18deantes, 4deantes, 1deantes}. Forensic radiology, through Computed Tomography (CT) scans, provides datasets particularly suited for training these algorithms, enabling objective, rapid, and highly reproducible analysis~\cite{degiorgio2022, 21deantes}. Post-mortem remains are subject to taphonomic processes that may alter anatomical structures and affect image interpretation~\cite{guareschi2023}. These alterations can increase image complexity and variability, while image degradation and noise are known to adversely affect the performance of deep learning classifiers in radiological imaging tasks~\cite{lin2026effects}.

Although recent studies have demonstrated the feasibility of deep learning for sex estimation using cranial and pelvic images~\cite{5deantes}, many remain restricted to specific populations or highly controlled acquisition conditions~\cite{18deantes}. Consequently, algorithms trained on these narrow datasets often fail to generalize to the nature of actual forensic casework. There remains a persistent need to validate these approaches using samples that reflect true forensic reality, including cadavers in varying stages of decomposition, morphological alterations due to trauma, and diverse age groups~\cite{morantorres2025}.

This work evaluates the robustness and feasibility of 
deep learning methods, enhanced by transfer learning, for the automated determination of biological sex. The distinctive feature of this work lies in its data source: CT scans of the pelvis and skull from a complex forensic collection of 141 cadavers autopsied at the Medical-Legal Institute of Goiânia-GO. The main contributions of this work are: (i) the validation of deep learning architectures on a diverse dataset of real-world forensic cases featuring significant anatomical disfigurements and trauma-related imaging artifacts; (ii) a comparative analysis of the discriminative power between pelvic and cranial two-dimensional (2D) projections; and (iii) an assessment of the generalization capabilities of transfer learning in the presence of forensic-specific image artifacts.


The rest of this paper is organized as follows: Section II reviews the related work in deep learning and forensic anthropology; Section III details the proposed methodology, including dataset characteristics and deep learning architecture; Section IV presents the experimental results and performance metrics; Section V provides a multidisciplinary discussion of the findings; and Section VI concludes the paper with final remarks and future directions.

\section{Related Works}
\label{sec:related_works}
The estimation of biological characteristics from human remains has been an essential field of study in forensic anthropology, especially for disaster victim identification (DVI). Historically, biological sex determination has relied on the observation of dimorphic morphological traits and established bone measurement metrics. While these classical methods, such as the Walker standards~\cite{b1} and the Phenice method~\cite{phenice1969} for cranial and pelvic assessment, respectively, show high accuracy under favorable conditions, their diagnostic efficacy is significantly reduced in the presence of sample degradation or advanced decomposition~\cite{b2}. Furthermore, these techniques are vulnerable to subjective visual bias and variability in observer capacity~\cite{b2, klales2012}. To overcome these limitations, the integration of CT with AI architectures has shifted the paradigm toward automated, highly objective new technologies for biological profiling~\cite{5deantes,1deantes}.

\subsection{Deep Learning in Skeletal Morphometrics}
Recent advancements in medical computer vision have established Convolutional Neural Networks (CNNs) as powerful tools for autonomously extracting hierarchical features from complex osseous structures~\cite{CNN_Melhor_Altern}. Early deep learning applications primarily targeted isolated skeletal elements with pronounced sexual dimorphism. For example, a study used the GoogLeNet Inception V4 architecture on pelvic CT reconstructions, achieving an AUC over 0.90 for highly dimorphic regions such as the greater sciatic notch and the ventral pubis, outperforming human experts \cite{18deantes}.

In a recent study \cite{5deantes}, a multi-task convolutional neural network (CNN) achieved 97\% accuracy in cranial sex estimation, significantly outperforming human experts, who attained 82\% accuracy using traditional morphoscopic assessment methods. Crucially, the deployment of Explainable AI (XAI) techniques, notably Gradient-weighted Class Activation Mapping (Grad-CAM), has validated these computational approaches~\cite{9093461}. Studies confirm that deep learning algorithms autonomously converge on established anthropological landmarks such as the glabella, supraorbital margin, and mental eminence, ensuring that model predictions are grounded in valid anatomical features rather than spurious imaging artifacts~\cite{5deantes,21deantes}. 

Recent benchmarking efforts demonstrate that the choice between modernized CNNs (e.g., ConvNeXt, EfficientNetV2) and Vision Transformers (ViTs) is context-dependent, as both architectures offer distinct advantages regarding data scale, efficiency, and generalization. A study~\cite{ms_benchmark2026} evaluated 23 AI architectures, including CNNs and Transformers, for sex classification from maxillary sinus radiographs revealed that while ViTs offer unprecedented global contextual awareness, modernized CNNs and real-time frameworks provide highly competitive accuracy with significantly lower computational overhead. Furthermore, YOLO architectures have been successfully deployed for sex prediction using highly localized skeletal regions, such as molar teeth, demonstrating their robustness in scenarios involving compromised or incomplete biological material~\cite{yolo_molar}.

\subsection{Data Scarcity and Dimensionality of Dataset}
A challenge in the development of robust forensic AI models is the chronic scarcity of large, annotated datasets derived from diverse, contemporary populations~\cite{BATOOL2025107661}. To mitigate severe overfitting in data-sparse environments, Transfer Learning (TL) has become a methodological imperative \cite{Weiss2016}. By initializing networks with weights pre-trained on massive, general-purpose visual databases (e.g., ImageNet~\cite{9deantes}), models can leverage established visual priors, allowing them to converge rapidly on specialized medical tasks even with limited sample sizes \cite{20deantes}.

Furthermore, processing high-resolution three-dimensional (3D) volumetric CT data directly with 3D-CNNs incurs increased computational overhead and exacerbates data demands. To optimize algorithmic efficiency without sacrificing diagnostic fidelity, researchers have increasingly adopted dimensionality reduction strategies. Landmark studies have demonstrated the viability of utilizing 2D orthogonal projections extracted from postmortem CT (PMCT) volumes~\cite{seo}. By applying transfer learning to 2D skull silhouettes, their models achieved up to 91.7\% accuracy using multi-angle voting systems. Simplified 2D representations successfully capture highly dimorphic outer contours, such as the superciliary arch and external occipital protuberance, without incurring the increased computational penalty of processing the internal volumetric architecture~\cite{seo}. This validates the shift toward comprehensive multi-view projections, a methodology continuously reinforced in modern forensic contexts, such as the rapid estimation of biological profiles directly from 2D PMCT scout views~\cite{pmct_scout_2026}.

Despite these advancements in dimensionality reduction, a critical gap remains in identifying frameworks that optimally balance low computational latency, parameter efficiency, and multi-scale feature sensitivity. Recent literature highlights the need to systematically assess the computational cost of modern medical imaging models, and to explore lightweight CNNs as efficient alternatives to heavy self-attention operations~\cite{mednext_2026}. Extensive benchmarking of contemporary models confirms that advanced architectures, such as the YOLO11 and YOLO26 families, excel in latency-constrained scenarios. By utilizing highly optimized backbones, these frameworks provide superior feature-extraction efficiency and exceptional parameter-to-accuracy ratios compared to heavier baselines~\cite{yolo26_benchmark_2026}. Therefore, benchmarking these real-time computational frameworks against modernized CNNs and ViTs is essential to establish scalable and robust pipelines for forensic deployment.


\section{Methods}
\subsection{Study Design and Sample}
\label{sec:guidelines}
This retrospective study was conducted using forensic data provided by the Aristoclides Teixeira Institute of Forensic Medicine (IMLAT), Goiânia, Goiás, Brazil. The dataset was collected and curated by forensic specialists from the institute and consisted of PMCT scans acquired as part of routine forensic investigations.

The total sample consisted of CT images from 143 cadavers examined at IMLAT, including 104 males and 39 females. This class imbalance reflects the stochastic distribution of forensic services and was addressed during the training phase through weighted loss functions and data augmentation to prevent majority-class bias. 
The sample covered a broad and heterogeneous age range.

\begin{figure}[t!]
    \centering
    \includegraphics[width=0.8\columnwidth]{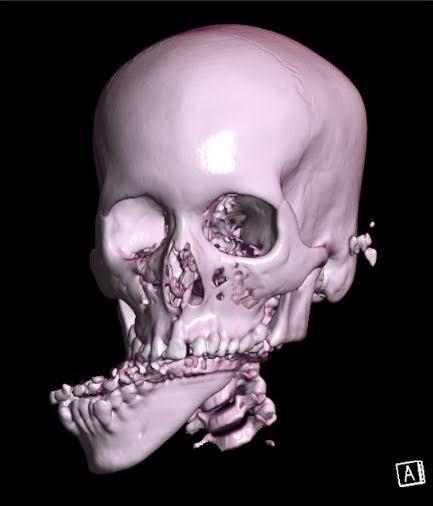}
    \caption{Reconstructed image presenting maxillary deformation due to trauma; key cranial landmarks remain identifiable.}
    \label{fig:cranio_trauma}
\end{figure}

\begin{figure}[t!]
    \centering
    \includegraphics[width=0.8\columnwidth]{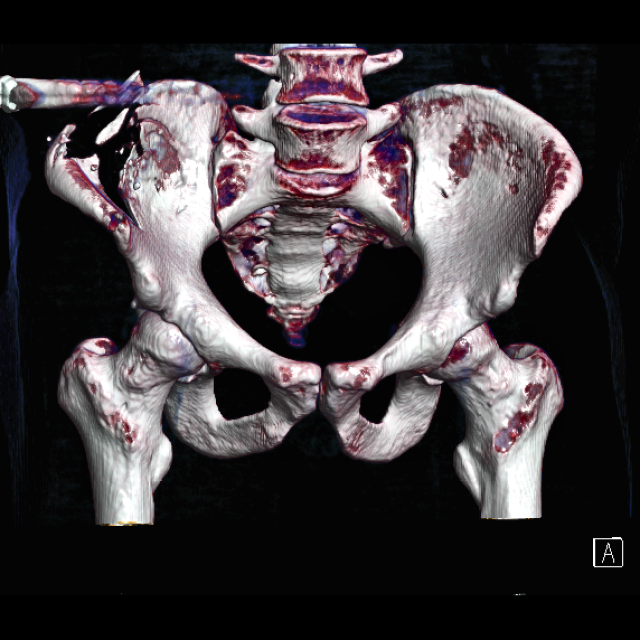}
    \caption{The iliac crest of the pelvis shows signs of impairment.}
    \label{fig:pelve_trauma}
\end{figure}

In order to train the model under forensic conditions as close as possible to real-world scenarios, strict exclusion criteria were intentionally avoided. Cases presenting significant anatomical disfigurements and imaging artifacts were therefore retained in the dataset (Fig.~\ref{fig:cranio_trauma} and Fig.~\ref{fig:pelve_trauma}). Only scans showing extensive bone destruction, virtually always associated with high-energy trauma, sufficient to prevent the identification of any morphological landmark necessary for classification were excluded.

\subsection{Data Acquisition and Preprocessing}

The raw data were obtained in the Digital Imaging and Communications in Medicine (DICOM) format \cite{6deantes}. Each examination consisted of a series of axial slices of the skull and pelvis, as well as 3D reconstructions of the same anatomical structures. The images were acquired using a scanner called "Siemens SOMATOM go.Now" with a slice thickness of 0.6 mm and no inter-slice gap. The acquisition protocol generated two volumetric blocks: one covering the skull and another including the pelvis.

Subsequently, three-dimensional reconstructions were generated via Volume Rendering Technique (VRT)~\cite{7deantes}. The use of 2D projections extracted from these volumes serves as a dimensionality-reduction strategy, mapping complex 3D spatial relationships into high-contrast 2D manifolds optimized for deep learning feature extraction. Regions of interest (skull and pelvis) were extracted at the workstation using the "syngo.via platform" (Siemens Healthineers), where non-osseous structures were removed via segmentation and windowing adjustments, ensuring the model relied exclusively on skeletal morphology. To ensure ethical compliance and forensic confidentiality, all data were anonymized on secure IMLAT servers prior to computational processing.

For this study, the angular visualization series simulating a 360° rotation of the structures was selected. Custom Python scripts, utilizing the pydicom library~\cite{8deantes}, were developed to automate the screening of these series based on the Body Part Examined metadata. Subsequently, each 2D frame composing the rotation was extracted and converted into PNG format.

Following the initial extraction, a rigorous data cleaning and quality control process was conducted for each subject. To ensure dataset integrity, we removed redundant directories containing duplicate folders for the same anatomical region. Furthermore, subjects with insufficient data, such as those with only three slices, were excluded, along with cases flagged by medical experts for clinical inconsistencies or imaging artifacts. This refinement process resulted in the final cohort of 141 cadavers described earlier, providing a validated, homogeneous foundation for subsequent analysis.

A critical methodological challenge was the absence of standardized metadata capable of unequivocally indicating the exact frontal (anterior) view of each anatomical structure, as CT table positioning varies due to post-mortem rigidity (rigor mortis), leading to 3D reconstructions with inconsistent rotational orientations.

To standardize the model inputs, a manual curation step was performed, during which a forensic specialist inspected each subject's image sequences and annotated a "pivot image" (\(\theta\)=0°), corresponding to the ideal frontal projection of each skull and pelvis. From this point, a rotational window of ±5 frames 
(approximately ±45°) was extracted, with 11 images per anatomical region. The final dataset had 3059 images, as shown in table \ref{tab:distribuicao_imagens}, slightly fewer than anticipated due to insufficient images for a few patients.

To establish a standard pose across all subjects, this view was chosen because it captures regions of interest for classification, such as the subpubic angle and pubic body in the pelvis, and the glabella, supraorbital margins, and mandibular angles in the skull, which are best exposed in anterior projection. We acknowledge that this choice excludes posterior landmarks such as the occipital protuberance and treat wider angular windows as future work.

The final dataset was organized into a hierarchical structure: Sex → Subject ID → Anatomical Region → 11 Images.

\begin{figure}[H]
    \centering
    \includegraphics[width=\columnwidth]{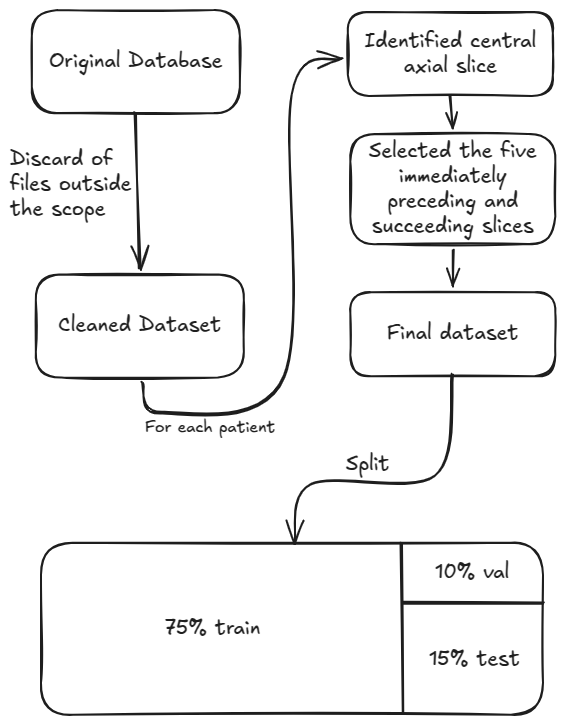}
    \caption{Schematic of the dataset construction from the original data.}
    \label{fig:esquematico}
\end{figure}

The process of building the final dataset is illustrated in Fig.~\ref{fig:esquematico}. To ensure an objective evaluation and prevent data leakage, the dataset split was strictly performed at the subject level. This means all images belonging to a single cadaver were assigned exclusively to one partition, ensuring no subject appeared simultaneously in the training and testing phases. Following this protocol, 75\% of the cadavers were allocated for training, 10\% for validation to monitor convergence, and 15\% were held out exclusively for the final testing phase. The number of images for each division is shown in table \ref{tab:distribuicao_imagens}.

This study was approved by the CEPTC (Education Coordination of the Scientific Technical Police) Ethics Committee under protocol \#202500016022484.

\begin{table}[H]
    \centering
    \caption{Distribution of images in the dataset}
    \label{tab:distribuicao_imagens}
    \begin{tabular}{|l|c|c|c|c|}
         \hline & \textbf{Train} & \textbf{Validation} & \textbf{Test} & \textbf{Total} \\ \hline
         Number of Images        & 2267           & 286                & 506            & 3059                \\ \hline
    \end{tabular}
\end{table}

\subsection{Baseline Architecture Comparison and Training Setup}
To evaluate the efficiency of pre-trained models to extract features from our dataset, we conducted a comparative analysis of seven distinct architectures: EfficientNetV2~\cite{efficientnetv2}, ConvNeXt~\cite{convnext2022}, ViT-B16~\cite{vit2020}, VGG16~\cite{vgg2014}, ResNet50~\cite{resnet2016}, YOLO11~\cite{ultralytics_yolo11}, and YOLO26~\cite{11deantes}. The models were trained to classify the images into four distinct categories: male skull, female skull, male pelvis, and female pelvis.

All architectures were trained using their default hyperparameters and optimizers to establish a standardized baseline. The training sessions were configured for 300 epochs, with an early stopping callback that halted training upon convergence to mitigate overfitting. As described in the subsequent section, YOLO26 had the best overall performance.

With that determinated, three  architectures from YOLO were also evaluated, namely Nano, Small, and Medium, which differ progressively in model complexity, i.e., in the number of layers and parameters ~\cite{wong2019yolonanohighlycompact,11deantes}. 

The YOLO26 models were trained and compared under two distinct classification scenarios:
\begin{itemize}
    \item \textbf{Binary classification:} prediction of biological sex only (Male/Female), independent of the anatomical region.
    
    \item \textbf{Quaternary classification:} joint prediction of biological sex and anatomical region (Male Skull, Female Skull, Male Pelvis, Female Pelvis).
\end{itemize}

\subsection{Data Augmentation}
To mitigate the impact of the limited sample size (141 cadavers) and the class imbalance (103 males / 38 females), data augmentation techniques were employed~\cite{10deantes}. 


Three different data augmentation approaches were evaluated \cite{ogara2019comparing}. The first approach, an custom offline augmentation, utilized a transformation pipeline implemented via the Albumentations library~\cite{10deantes}. To isolate the effects of this offline process, the model’s native online augmentation functions were disabled. Each original image in the dataset was used to generate four additional augmented images, expanding the dataset by a factor of five, from 3059, as seen in Table \ref{tab:distribuicao_imagens}, to 15295 images.

During each generation step, a sequence of transformations was applied, each with an associated probability of occurrence. The transformations included random resized cropping (RandomResizedCrop), mild rotations (up to 30 degrees), color shifts (RGBShift), slight image compression (ImageCompression), noise injection (GaussNoise), horizontal flipping, and sharpening (Sharpen). These transformations were selected with the objective of preserving the morphological integrity of skeletal structures. Transformations that could introduce structural distortions (e.g., grid-based deformations) were intentionally avoided to ensure that the model learned from realistic anatomical patterns.

The second approach, YOLO Default Augmentation~\cite{12deantes}, relied exclusively on native data augmentation strategies provided by the Ultralytics~\cite{11deantes} implementation of YOLO. In this configuration, all default parameters of the framework standard online augmentation were maintained. This approach dynamically applies a set of transformations during training, including mosaic augmentation~\cite{13deantes}, translation, scaling, and horizontal flipping.

The third approach, Hybrid Augmentation, combined the two previously described strategies. In this configuration, the dataset was first expanded using the custom augmentation pipeline and subsequently subjected to the native augmentation mechanisms applied dynamically during model training.

\subsection{Validation Protocol and Metrics}

All performance results (e.g., accuracy, confusion matrix) reported in the Results section refer exclusively to the models' performance on the test set (15\%), thus providing the most reliable estimate of their real-world applicability.

To produce a single diagnosis per subject, the model's outputs across the 11 angular projections of each anatomical structure were aggregated via soft voting. Rather than counting per-image predicted labels (majority voting), the softmax probability vectors of the 11 projections were averaged, and the patient-level class was determined as defined in Eq.~\ref{eq:soft_voting}.

\begin{equation}
\label{eq:soft_voting}
y_{\text{subject}} =
\arg\max_{c}
\left(
\frac{1}{K}
\sum_{k=1}^{K}
p_k(c)
\right),
\quad K = 11
\end{equation}

This probability-level fusion preserves each projection's confidence, allowing high-certainty views to outweigh ambiguous ones, and mitigates isolated misclassifications caused by unfavorable viewing angles or localized imaging artifacts.

To further evaluate the robustness of the dataset and the generalization capability of the architecture, a complementary 5-fold cross-validation strategy~\cite{k-fold} was implemented. This additional approach ensures that the dataset's consistency is verified across different training scenarios, confirming its usability beyond a single random split. For these validation experiments, the dataset was partitioned into five distinct folds; in each iteration, four folds were used for training while the remaining one served as the validation set. All folds were trained with a batch size of 64 for a maximum of 100 epochs, and the weights were saved at the \textit{best epoch} based on validation performance to prevent overfitting.

Standard classification metrics, accuracy, precision, recall (sensitivity) and F1-score, were employed to evaluate model performance~\cite{15deantes}. While accuracy provides an overall performance overview, the F1-score was prioritized to account for class imbalance, as it combines precision and recall into a single harmonic mean. To enable a multidimensional assessment, precision, recall, and F1-score were computed both per class and as aggregate (macro-averaged) measures, alongside the overall accuracy, in each classification scenario. Evaluating these indicators across both the binary and quaternary settings provided a comprehensive basis for assessing the effectiveness of the proposed augmentation strategies.

Finally, to ensure the interpretability of the deep learning models and verify that their predictions are grounded in valid anatomical features, Grad-CAM Class Activation Maps (heatmaps)~\cite{grad-cam} were generated to visualize the network's spatial attention~\cite{wang2023explainable}. This technique was chosen because it has been validated in the literature as a method for skeletal sex-estimation models, where it has been used to confirm that network attention aligns with recognized dimorphic landmarks.~\cite{9093461, 16deantes}





\subsection{Hyperparameter Optimization via Genetic Algorithm}

To maximize the performance and generalization of the models, hyperparameter optimization was performed using a Genetic Algorithm (GA). Following the baseline comparison, only the best-performing model was subjected to the optimization procedure. 
This approach is particularly advantageous for exploring complex, multidimensional search spaces, enabling the discovery of optimal parameters in a very fast, computationally efficient manner~\cite{algoritimo}. 

The optimization process was conducted using the native automatic tuning tool, called auto-tuning, from the Ultralytics library. The algorithm was executed over 100 iterations. The \textit{fitness function} evaluated the success of each set of hyperparameters based on a weighted combination of classification metrics.

The algorithm's search space encompassed both the neural network's learning dynamics and the data augmentation settings. After the evolutionary process, the optimal set of hyperparameters was defined and applied in the definitive training of the models. Furthermore, the GA fine-tuned the image transformations to better suit the tomographic dataset.

\subsection{Model Refinement and Imbalance Compensation}

Upon determining the optimal hyperparameters through the optimization phase, the final model was trained using a specialized configuration to account for dataset characteristics. A primary challenge identified was the significant class imbalance between male and female patients, with a marked over-representation of the former~\cite{Imbalance}. To prevent the model from developing a classification bias toward the majority class, we implemented a Weighted Cross-Entropy Loss.

\subsubsection{Weighted Loss Formulation}
In standard training, the model may achieve high overall accuracy by prioritizing the majority class, leading to poor generalization for the minority group. To mitigate this, we utilized an inverse-frequency weighting scheme. The weight $w_i$ for each class $i$ is calculated as defined in Eq.~\ref{eq:w_i}.

\begin{equation}\label{eq:w_i}
w_i = \frac{N}{C \cdot n_i}
\end{equation}

where $N$ represents the total sample size, $C$ denotes the number of categories, and $n_i$ is the specific sample count for class $i$. This adjustment ensures that the loss contribution from female patient samples is proportionally higher, effectively penalizing errors on the minority class more severely.

\subsubsection{Integration and Backpropagation}
The custom loss function was injected into the training architecture via an \textit{on\_train\_start} callback. This integration is critical during backpropagation, as scaling the loss amplifies the gradients for minority-class samples. Consequently, the optimizer performs more aggressive weight updates when the model misclassifies the underrepresented group, ensuring a balanced learning signal and equitable performance across both sexes.

\subsection{Multi-Criteria Model Selection: TOPSIS and VIKOR}
\label{sec:topsisvikor}
To consolidate the ranking and eliminate interpretation biases, two complementary methods were used. First, the Technique for Order of Preference by Similarity to Ideal Solution (TOPSIS) method was applied~\cite{TOPSIS}. The multicriteria decision analysis follows a structured mathematical procedure consisting of the following steps:

\begin{enumerate}
    \item \textbf{Establishment and Normalization of the Decision Matrix:} Let $X = [x_{ij}]_{m \times n}$ be the decision matrix, where $x_{ij}$ represents the performance of the $i$-th model ($i = 1, \dots, m$) with respect to the $j$-th metric ($j = 1, \dots, n$). To eliminate scale disparities, the matrix is normalized using vector normalization, resulting in $R = [r_{ij}]_{m \times n}$ as defined by Eq.~\ref{eq:eq_a}.
    \begin{equation}\label{eq:eq_a}
    r_{ij} = \frac{x_{ij}}{\sqrt{\sum_{i=1}^{m} x_{ij}^2}}
    \end{equation}

    \item \textbf{Application of Criteria Weights:} The weighted normalized decision matrix $V = [v_{ij}]_{m \times n}$ is calculated by multiplying the normalized values by their respective weights $w_j$, as defined by Eq.~\ref{eq:eq_b}.
    \begin{equation}\label{eq:eq_b}
    v_{ij} = w_j \cdot r_{ij}
    \end{equation}
    where $\sum_{j=1}^{n} w_j = 1$. In this study, equal weights were assigned to all evaluation metrics ($w_j = 1/n$).

    \item \textbf{Determination of Ideal Solutions:} The positive ideal solution ($V^+$) and the negative ideal solution ($V^-$) are identified. Since all evaluated metrics in this study are benefit criteria (where higher values denote superior performance), they are defined as by Eq.~\ref{eq:eq_c}.
    \begin{equation}\label{eq:eq_c}
    V^+ = (v_1^+, v_2^+, \dots, v_n^+) = (\max_{i} v_{ij} \mid j = 1, \dots, n)
    \end{equation}
    \begin{equation}
    V^- = (v_1^-, v_2^-, \dots, v_n^-) = (\min_{i} v_{ij} \mid j = 1, \dots, n)
    \end{equation}

    \item \textbf{Calculation of Separation Measures:} The geometric Euclidean distances of each alternative from the positive ideal solution ($D_i^+$) and the negative ideal solution ($D_i^-$) are computed as defined in Eq.\ref{eq:eq_d}.
    \begin{equation}\label{eq:eq_d}
    D_i^+ = \sqrt{\sum_{j=1}^{n} (v_{ij} - v_j^+)^2}
    \end{equation}
    \begin{equation}
    D_i^- = \sqrt{\sum_{j=1}^{n} (v_{ij} - v_j^-)^2}
    \end{equation}

    \item \textbf{Computation of the Relative Proximity Coefficient:} The final performance score ($C_i$) for each model is determined by Eq.~\ref{eq:topsis}.
    \begin{equation}
    \label{eq:topsis}
    C_i = \frac{D_i^-}{D_i^+ + D_i^-}
    \end{equation}
    where $0 \le C_i \le 1$. Alternatives with higher $C_i$ values rank closer to the ideal solution.
\end{enumerate}

Subsequently, the VIKOR method~\cite{VIKOR}, employing objective weights derived from Shannon Entropy~\cite{Entropia_Shannon}. The application of Entropy eliminates subjective biases by assigning weights based on each evaluation metric's intrinsic statistical discrimination power.

Let $X = [x_{ij}]_{m \times n}$ be the decision matrix with $m$ models and $n$ metrics. Initially, the matrix is linearly normalized such that $p_{ij} = x_{ij} / \sum_{i=1}^{m} x_{ij}$. The entropy measure $e_j$ for each criterion $j$ is calculated as Eq.~\ref{eq:eq_f}.
\begin{equation}\label{eq:eq_f}
e_j = -k \sum_{i=1}^{m} p_{ij} \ln(p_{ij})
\end{equation}
where $k = 1/\ln(m)$ is a constant that guarantees $0 \le e_j \le 1$. The objective weight $w_j$ is then derived from the degree of diversification $d_j = 1 - e_j$ and normalized across all criteria as defined by Eq.\ref{eq:eq_g}.
\begin{equation}\label{eq:eq_g}
w_j = \frac{d_j}{\sum_{j=1}^{n} d_j}
\end{equation}

With the weights established, the VIKOR algorithm evaluates the models based on their proximity to the ideal solution. Let $f_j^+ = \max_i x_{ij}$ and $f_j^- = \min_i x_{ij}$ define the best and worst values achieved for each benefit criterion. The method computes the group utility index ($S_i$, representing the majority agreement) and the individual regret index ($R_i$, representing the maximum disagreement) for each alternative, as given by Eq.~\ref{eq:eq_h}.
\begin{equation}\label{eq:eq_h}
S_i = \sum_{j=1}^{n} w_j \frac{(f_j^+ - x_{ij})}{(f_j^+ - f_j^-)} 
\end{equation}
\begin{equation}
R_i = \max_j \left[ w_j \frac{(f_j^+ - x_{ij})}{(f_j^+ - f_j^-)} \right]
\end{equation}

The final multi-criteria compromise ranking index $Q_i$ is determined by blending the two components as defined in Eq.~\ref{eq:eq_i}.
\begin{equation}\label{eq:eq_i}
Q_i = v \left( \frac{S_i - \min_k S_k}{\max_k S_k - \min_k S_k} \right) + (1 - v) \left( \frac{R_i - \min_k R_k}{\max_k R_k - \min_k R_k} \right)
\end{equation}

where $v$ represents the weight of the "majority of criteria" strategy. In this formulation, $v = 0.5$ was adopted, ensuring an equitable consensus between average performance and individual dimension stability. In the VIKOR framework, alternatives with lower $Q_i$ values indicate a superior compromise solution. 


\section{Results}

\subsection{Baseline Architecture Comparison}
The performance metrics for baseline models evaluated are summarized in Table~\ref{tab:model_performance}. Several architectures achieved strong performance, exceeding 80\% in Accuracy. In contrast, older or more complex architectures exhibited poor convergence on our limited dataset. YOLO26 emerged as the top-performing model, consistently surpassing 89\% across all evaluated metrics. Consequently, YOLO26 was selected as the foundational architecture for the remainder of this research.

Although Table~\ref{tab:model_performance} reflects each architecture's performance under default settings rather than its fully tuned ceiling, the clear lead of YOLO26 across all metrics motivated its selection for the subsequent optimization stages. The same hyperparameter and augmentation optimization were not applied to the remaining architectures.

\begin{table}[H]
\centering
\caption{Performance Metrics Comparison of Baseline Deep Learning Models}
\label{tab:model_performance}
\ra{1.3}
\begin{tabular}{|l|c|c|c|c|}
\hline
\textbf{Model} & \textbf{Accuracy} & \textbf{Recall} & \textbf{F1-score} & \textbf{Precision} \\ \hline
ConvNeXt-Tiny   & 83.00\%           & 78.65\%         & 79.20\%           & 80.05\%            \\ \hline
EfficientNetV2 & 80.63\%           & 75.85\%         & 75.54\%           & 76.91\%            \\ \hline
ViTB-16         & 71.14\%           & 61.18\%         & 57.46\%           & 66.17\%            \\ \hline
VGG16          & 64.62\%           & 63.62\%         & 59.89\%           & 65.88\%            \\ \hline
ResNet50       & 77.27\%           & 69.05\%         & 68.62\%           & 71.73\%            \\ \hline
YOLO11        & 88.34\%           & 88.45\%         & 84.96\%           & 83.04\%            \\ \hline
YOLO26         & 91.30\%           & 89.42\%         & 89.89\%           & 90.46\%            \\ \hline
\end{tabular}
\end{table}

\subsection{Evaluation Metrics}

Two classification strategies were tested: binary and quaternary. For each strategy, nano, small, and medium YOLO26 models were trained with the different augmentation approaches mentioned before, including YOLO default augmentation, Custom augmentation, and a combination of both.

\begin{figure}[H]
\centering
\includegraphics[scale=0.3]{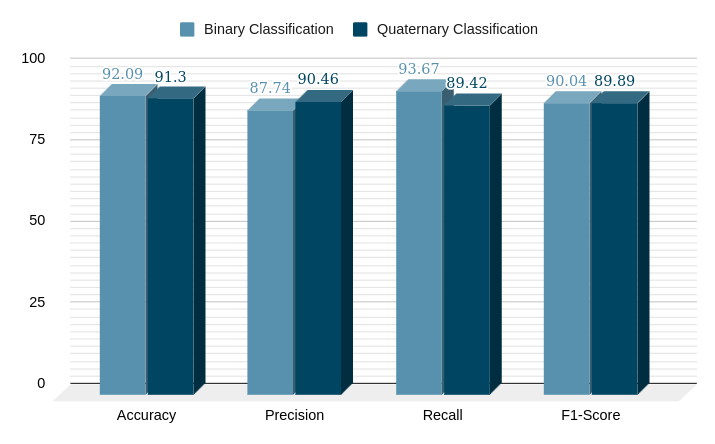}
\caption{\textbf{Evaluation metrics of the models that presented the best statistics based on the dataset test split.}}
\label{fig5}
\end{figure}

For a comparative analysis of the results, the models with the best overall accuracy in each strategy were selected, specifically the Medium model with binary classification, which used hybrid data augmentation (customized + YOLO), and the Nano model with quaternary classification, which employed only standard YOLO data augmentation, as shown in Fig.~\ref{fig5}.
 
\begin{table}[H]
\centering
\caption{Comparison of Precision Extremes: Binary vs. Quaternary}
\label{tab:precision_extremes}
\ra{1.3}
\resizebox{\columnwidth}{!}{
\begin{tabular}{|l|c|c|c|}
\hline
\textbf{Model} & \textbf{Overall Precision} & \textbf{Lowest Class Precision} & \textbf{Highest Class Precision} \\ \hline
Binary & 87.74\% & 76.62\% & 98.86\% \\ \hline
Quaternary & 90.46\% & 88.31\% & 94.32\% \\ \hline
\end{tabular}%
}
\end{table}

Although the quaternary approach results in slightly lower performance than the binary one, its results remain robust and competitive, standing out fundamentally for superior class balance, as shown in Table \ref{tab:precision_extremes}. While the binary approach suffers from pronounced imbalance, in which performance for the female sex is significantly hindered by the limited number of samples, the quaternary strategy mitigates this disparity.

By separating the skull and pelvis into distinct categories, the network effectively reduces intra-class variance and isolates specific dimorphic features, avoiding the imposition of a universal biological pattern that would penalize the minority class. Consequently, the quaternary classification framework using the YOLO26 architecture was established as the definitive strategy, prioritizing diagnostic robustness and metric equity between sexes.

 As previously outlined, the quaternary strategy was implemented across distinct model scales (Nano, Small, and Medium) and subjected to varying data augmentation policies, specifically exploring native YOLO augmentations, a custom-designed pipeline, and a combined approach (YOLO + Custom). Although preliminary descriptive analysis indicated that the Nano architecture with default YOLO augmentation achieved the highest global accuracy, relying solely on a single metric is insufficient to definitively determine the optimal configuration among all trained alternatives. To consolidate the comparison into an objective ranking free of interpretation bias, the two multi-criteria decision-making methods described in Section~\ref{sec:topsisvikor} were applied over a decision matrix comprising 16 performance criteria, which integrated class-specific and overall metrics for Precision, Recall, and F1-Score, alongside global Accuracy. The first, TOPSIS, ranks each configuration by its proximity to the ideal solution, so that higher proximity coefficients ($C_i$) indicate superior overall performance.

The empirical results of the TOPSIS evaluation are summarized in Table~\ref{tab:topsis_results}.

\begin{table}[htbp]
\centering
\caption{TOPSIS Ranking and Proximity Coefficients for the Evaluated Models.}
\label{tab:topsis_results}
\ra{1.3}
\begin{tabular}{|l|c|c|}
\hline
\textbf{Model - Augmentation} & \textbf{Score ($C_i$)} & \textbf{Ranking} \\ \hline
Nano - YOLO                  & 0.7912                 & 1               \\ \hline
Small - YOLO                 & 0.7346                 & 2               \\ \hline
Nano -  Custom               & 0.6698                 & 3               \\ \hline
Small - YOLO + Custom        & 0.6520                 & 4               \\ \hline
Nano - YOLO + Custom         & 0.5445                 & 5               \\ \hline
Medium - Custom              & 0.5440                 & 6               \\ \hline
Small - Custom               & 0.4942                 & 7               \\ \hline
Medium - YOLO                & 0.3816                 & 8               \\ \hline
Medium - YOLO + Custom       & 0.2022                 & 9               \\ \hline
\end{tabular}
\end{table}

The results identified the \textit{Nano} model with \textit{YOLO Default Augmentation} with the highest $C_i$ value, proving that this configuration not only converges toward the theoretical ideal solution in terms of absolute performance but also demonstrates resilience by remaining consistently distant from the lower performance limits recorded in the sample set.

To confirm this ranking under a different aggregation logic, the VIKOR method was then applied, in which lower compromise indices ($Q_i$) indicate a better balance between group utility and individual regret. The empirical results of the Shannon Entropy–VIKOR evaluation are detailed in Table~\ref{tab:vikor_results}.

\begin{table}[htbp]
\centering
\caption{VIKOR Compromise Indexes and Final Robustness Ranking.}
\label{tab:vikor_results}
\resizebox{\columnwidth}{!}{
\begin{tabular}{|l|c|c|c|c|}
\hline
\textbf{Model - Augmentation} & \textbf{Utility ($S_i$)} & \textbf{Regret ($R_i$)} & \textbf{Index ($Q_i$)} & \textbf{Ranking} \\ \hline
Nano - YOLO                  & 0.2173                   & 0.0614                 & 0.0000                 & 1               \\ \hline
Small - YOLO                 & 0.2644                   & 0.0768                 & 0.0887                 & 2               \\ \hline
Small - YOLO + Custom        & 0.3205                   & 0.1083                 & 0.2238                 & 3               \\ \hline
Nano - Custom                & 0.4860                   & 0.1340                 & 0.4286                 & 4               \\ \hline
Medium - Custom              & 0.5983                   & 0.1556                 & 0.5518                 & 5               \\ \hline
Small - Custom               & 0.6133                   & 0.1873                 & 0.6277                 & 6               \\ \hline
Nano  - YOLO + Custom        & 0.6407                   & 0.1873                 & 0.6419                 & 7               \\ \hline
Medium - YOLO                & 0.7290                   & 0.1989                 & 0.7186                 & 8               \\ \hline
Medium - YOLO + Custom       & 0.9860                   & 0.2732                 & 1.0000                 & 9               \\ \hline
\end{tabular}
}
\end{table}

The VIKOR index confirmed that the \textit{Nano} with \textit{YOLO Default Augmentation} model represents the optimal consensus solution, achieving the absolute minimum score ($Q_i = 0.0000$). This proves that this configuration not only leads to aggregated efficiency but also effectively mitigates risks of underperformance in critical sub-dimensions (classes). Both TOPSIS and VIKOR identified the Nano model with YOLO Default Augmentation as the top-ranked configuration, and although the two methods disagreed on the intermediate ordering, their agreement on the optimal solution supports its selection.

\subsection{Optimized Hyperparameters}

To enhance classification performance and ensure robust generalization, a hyperparameter tuning process was executed over 100 iterations. The optimization aimed to maximize a holistic fitness metric based on validation accuracy and loss. The peak performance was recorded at iteration 31, achieving a minimum validation loss of 0.0182, and an overall fitness score of 0.99825. This figure is the optimizer's selection criterion, computed on the validation split used to guide the search, and therefore represents an optimistically biased upper bound rather than an estimate of generalization. The unbiased performance of the selected configuration is reported on the held-out test set in Section~\ref{sec:finalmodel}.

The resulting optimized hyperparameters, encompassing optimization rates, regularization coefficients, and data augmentation magnitudes, are detailed in Table~\ref{tab:hyperparameters}.
\begin{table}[h]
\centering
\caption{Optimized Hyperparameters for the Final Classification Model.}
\label{tab:hyperparameters}
\ra{1.3}
\small
\begin{tabular}{|l|c|}
\hline
\textbf{Hyperparameter} & \textbf{Value} \\ \hline
Initial Learning Rate ($lr_0$) & 0.00808 \\ \hline
Final Learning Rate ($lr_f$) & 0.01990 \\ \hline
Momentum & 0.88458 \\ \hline
Weight Decay & 0.00083 \\ \hline
Warmup Epochs & 4.24701 \\ \hline
Warmup Momentum & 0.86765 \\ \hline
Box Loss Gain ($box$) & 7.32367 \\ \hline
Class Loss Gain ($cls$) & 0.66047 \\ \hline
Distribution Focal Loss Gain ($dfl$) & 1.60062 \\ \hline
HSV-Hue Augmentation ($hsv_h$) & 0.03474 \\ \hline
HSV-Saturation Augmentation ($hsv_s$) & 0.90000 \\ \hline
HSV-Value Augmentation ($hsv_v$) & 0.42337 \\ \hline
Rotation (degrees) & 0.01100 \\ \hline
Translation (fraction) & 0.05942 \\ \hline
Scaling (gain) & 0.60644 \\ \hline
Shear (degrees) & 0.00021 \\ \hline
Perspective & 0.00074 \\ \hline
Flip Upside-Down (probability) & 0.00053 \\ \hline
Flip Left-Right (probability) & 0.44380 \\ \hline
BGR Channel Mix & 0.01754 \\ \hline
Mosaic Augmentation & 1.00000 \\ \hline
Mixup Augmentation & 0.00110 \\ \hline
Cutmix Augmentation & 0.00167 \\ \hline
Copy-Paste Augmentation & 0.00049 \\ \hline
Close Mosaic (last epochs) & 8 \\ \hline
\end{tabular}
\end{table}

\subsection{Cross-Validation and Performance Analysis}
The results across all five validation folds demonstrate high consistency in model performance, as summarized in Table~\ref{tab:kfold_results}.

\begin{table}[htbp]
\centering
\caption{5-Fold Cross-Validation Performance Summary}
\label{tab:kfold_results}
\ra{1.3}
\begin{tabular}{|l|c|c|}
\hline
\textbf{Fold} & \textbf{Top-1 Accuracy} & \textbf{Best Epoch} \\ \hline
1             & 93.47\%                 & 80                  \\ \hline
2             & 84.53\%                 & 15                  \\ \hline
3             & 96.73\%                 & 45                  \\ \hline
4             & 86.10\%                 & 9                   \\ \hline
5             & 93.81\%                 & 47                  \\ \hline
\textbf{Mean} & \textbf{90.93\%}        & --                  \\ \hline
\end{tabular}
\end{table}

The models evaluated within the cross-validation framework achieved a mean Top-1 accuracy of 90.93\%. Although variance was observed regarding the optimal stopping point, with the best epoch ranging from 9 to 80, the consistently high Top-1 scores across all subsets confirm the robustness of the extracted features and the suitability of the dataset for the classification task. This variance in convergence speed suggests that certain data folds presented features that facilitated faster optimization, while the overall stable accuracy underlines the reliability of the architectural approach.

\subsection{Final Model Performance and Evaluation}
\label{sec:finalmodel}
The final model was trained using the hyperparameters listed in Table \ref{tab:hyperparameters}. To ensure methodological rigor and avoid data leakage, the dataset was partitioned at the subject level, meaning all images from a single individual were kept entirely within the same fold. Computational metrics were aggregated at two distinct operational granularities, image-level analysis and patient-level diagnostic consensus. 

At the image level, each of the 11 views on every anatomical region was evaluated independently. At this granularity, the model achieved an overall mean top-1 accuracy of 94.89\%, with a macro-averaged precision, recall, and F1-score of 95.62\%, 92.90\%, and 94.06\%, respectively. Conversely, the patient-level evaluation aggregated the predictions from the 11 projections per subject using a soft voting scheme, effectively reconstructing the holistic diagnostic approach used in forensic anthropology. This consensus strategy successfully mitigated isolated image-level misclassifications, robustly driving the overall patient-level mean accuracy up to 95.65\%, with a macro-averaged precision of 97.22\%, recall of 92.85\%, and F1-score of 94.36\%.

Table~\ref{tab:model_performances} details the granular, class-specific performance metrics across both evaluation scales, highlighting the precision, recall, and F1-score for each anatomical and biological sex category.

\begin{table}[htbp]
\centering
\caption{Detailed Cross-Validation Performance Metrics at Image and Patient Granularity Levels.}
\label{tab:model_performances}
\ra{1.3}
\small
\begin{tabular}{|l|l|c|c|}
\hline
\textbf{Anatomical Class} & \textbf{Metric} & \textbf{Image (\%)} & \textbf{Patient (\%)} \\ \hline
\multirow{3}{*}{Female Skull} 
 & Precision & 96.05  & 100.00 \\ \cline{2-4}
 & Recall    & 82.95  & 71.43  \\ \cline{2-4}
 & F1-Score  & 89.02  & 83.33  \\ \hline
\multirow{3}{*}{Male Skull} 
 & Precision & 92.02  & 88.89  \\ \cline{2-4}
 & Recall    & 98.30  & 100.00 \\ \cline{2-4}
 & F1-Score  & 95.05  & 94.12  \\ \hline
\multirow{3}{*}{Female Pelvis} 
 & Precision & 98.77  & 100.00 \\ \cline{2-4}
 & Recall    & 90.91  & 100.00 \\ \cline{2-4}
 & F1-Score  & 94.67  & 100.00 \\ \hline
\multirow{3}{*}{Male Pelvis} 
 & Precision & 95.63  & 100.00 \\ \cline{2-4}
 & Recall    & 99.43  & 100.00 \\ \cline{2-4}
 & F1-Score  & 97.49  & 100.00 \\ \hline \hline
\multirow{3}{*}{\textit{Macro Average}} 
 & Precision & 95.62  & 97.22  \\ \cline{2-4}
 & Recall    & 92.90  & 92.86  \\ \cline{2-4}
 & F1-Score  & 94.06  & 94.36  \\ \hline \hline
\multicolumn{2}{|l|}{\textbf{Overall Accuracy}} & \textbf{94.89} & \textbf{95.65} \\ \hline
\end{tabular}
\end{table}

A comparative analysis of anatomical regions reveals that the pelvic structures and male cranial structures achieved exceptionally high classification metrics, with the male skull, female pelvis, and male pelvis achieving a recall of 100.00\% at the patient-level diagnostic scale. At this aggregated scale, the model proved highly reliable for these classes, maintaining a patient-level overall accuracy of 95.65\%. 

Conversely, female cranial structures presented a distinct challenge. At the image level, the model achieved outstanding precision for the female skull (96.05\%), indicating very few false positives, but a lower recall of 82.95\%. Whereas patient-level soft voting was perfect for both pelvis classes and improved male skull recall, but reduced the female skull recall to 71.43\%. This inversion indicates that the residual image level errors were not randomly distributed across projections, in which case averaging the softmax outputs would have canceled them out, but were concentrated within a small subset of female subjects: for these cases, the model misclassified the majority of the 11 projections as male, so the averaged probability reinforced the incorrect label rather than correcting it. Because the affected skulls migrated into the male-skull predictions, the same errors reduced male-skull precision from 92.02\% to 88.89\% at the patient level, a complementary signature of the identical misclassifications viewed from the male class. The perfect female skull precision (100\%) confirms this manifested purely as false negatives for borderline female skulls, never as false positives. Given the small female skull test set, this figure is highly sensitive to individual cases and should be interpreted with that limitation in mind.

\subsection{Confusion Matrix}

To better understand the model's behavior at the image level, the confusion matrix constitutes an effective visual tool to interpret the specific patterns of correct and incorrect predictions across the evaluated projections (image level). 

\begin{figure}[H]
\centering
\includegraphics[width=1.05\columnwidth]{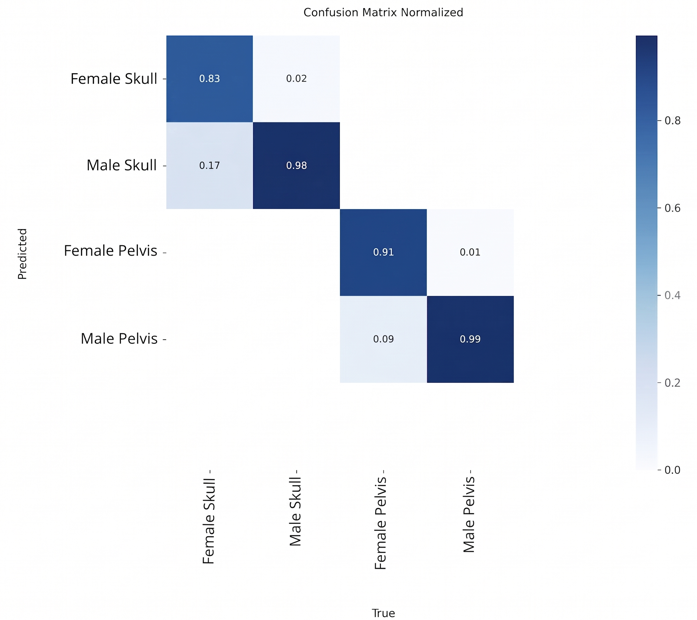}
\caption{\textbf{Normalized confusion matrix for the definitive 4-class model, representing the proportion of correct and incorrect predictions.}}
\label{fig:matriz_conf}
\end{figure}

For the quaternary approach, the normalized confusion matrix (Fig.~\ref{fig:matriz_conf}) provides a detailed analysis of the model's hits and misses, with the true and predicted axes divided into female skull, male skull, female pelvis, and male pelvis categories. It is worth noting that while the percentages displayed in the figure are rounded for visualization purposes, the exact values are reported in the text. The analysis of the matrix reveals good classification performance for the pelvic region, with 90.91\% of true female pelvis cases and 99.43\% of true male pelvis cases correctly classified. Similarly, the model demonstrated robust performance in the cranial region, particularly for male specimens, accurately identifying 98.30\% of true male skulls and 82.95\% of true female skulls. Furthermore, an anatomical error analysis indicates a consistent pattern within the limits of the available test set, establishing that misclassifications were strictly confined to sex differentiation within the same anatomical region, with no observed misclassification between skull and pelvis structures. Specifically, within the skull category, 17.05\% of the true female samples were incorrectly predicted as male, whereas only 1.70\% of the true male samples were misclassified as female. This trend was also reflected in the pelvis category, where 9.09\% of the true female samples were assigned to the male class, alongside a minimal 0.57\% error rate of true male samples classified as female.

The error distribution reveals a specific pattern where the proportion of true female structures incorrectly classified as male (17.05\% for skulls and 9.09\% for pelvises) is significantly higher than the reverse error rate (1.70\% and 0.57\%, respectively). This asymmetry indicates that, despite using the Weighted Cross-Entropy Loss to address class imbalance, the imbalance in the number of male samples (the majority class) continues to heavily influence the model's feature representation. Since the network is exposed to a wider variety of male anatomical structures during training, its internal boundaries remain intrinsically biased. Consequently, when confronted with ambiguous or borderline female specimens, the model falls back on the prevailing data distribution, resulting in a higher false-negative rate for female categories.

\subsection{Heatmaps}

\begin{figure}[h!]
\centering
\includegraphics[width=0.8\columnwidth]{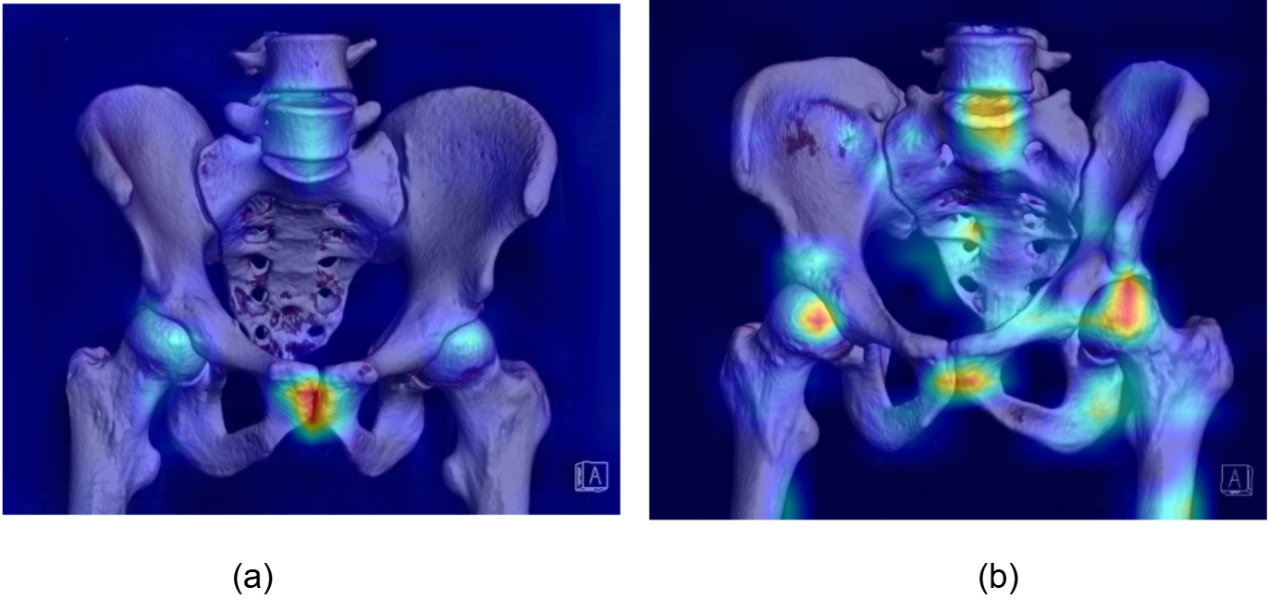}
\caption{\textbf{Male -- (a) Symphysis and subpubic angle. (b) Symphysis, subpubic angle, femoral head, and sacral promontory.}}
\label{fig7}
\end{figure}

\begin{figure}[h!]
\centering
\includegraphics[width=0.5\columnwidth]{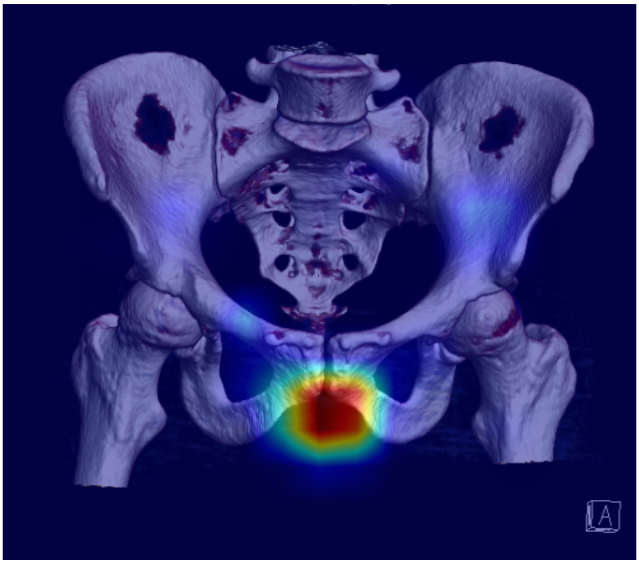}
\caption{\textbf{Female -- Subpubic angle.}}
\label{fig8}
\end{figure}

\begin{figure}[h!]
\centering
\includegraphics[width=0.8\columnwidth]{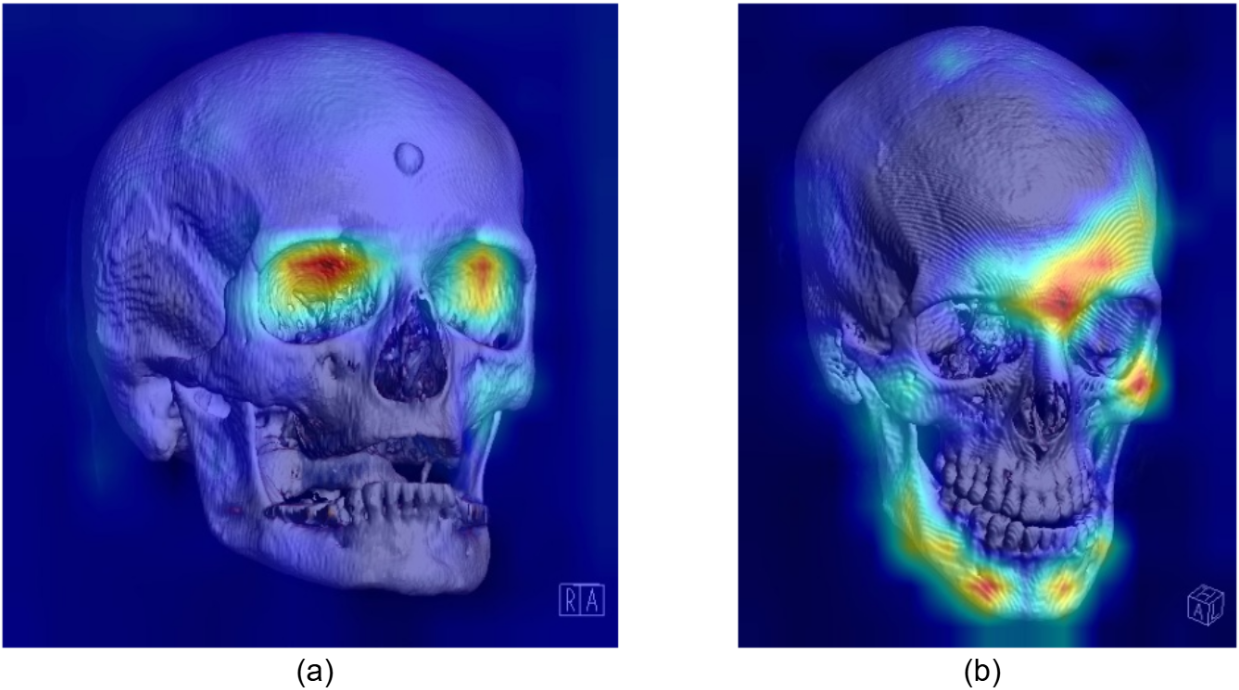}
\caption{\textbf{Female -- (a) Supraorbital margin and orbital cavity. (b) Mandibular ramus and angle, frontonasal angle, and glabella.}}
\label{fig9}
\end{figure}

\begin{figure}[h!]
\centering
\includegraphics[width=0.8\columnwidth]{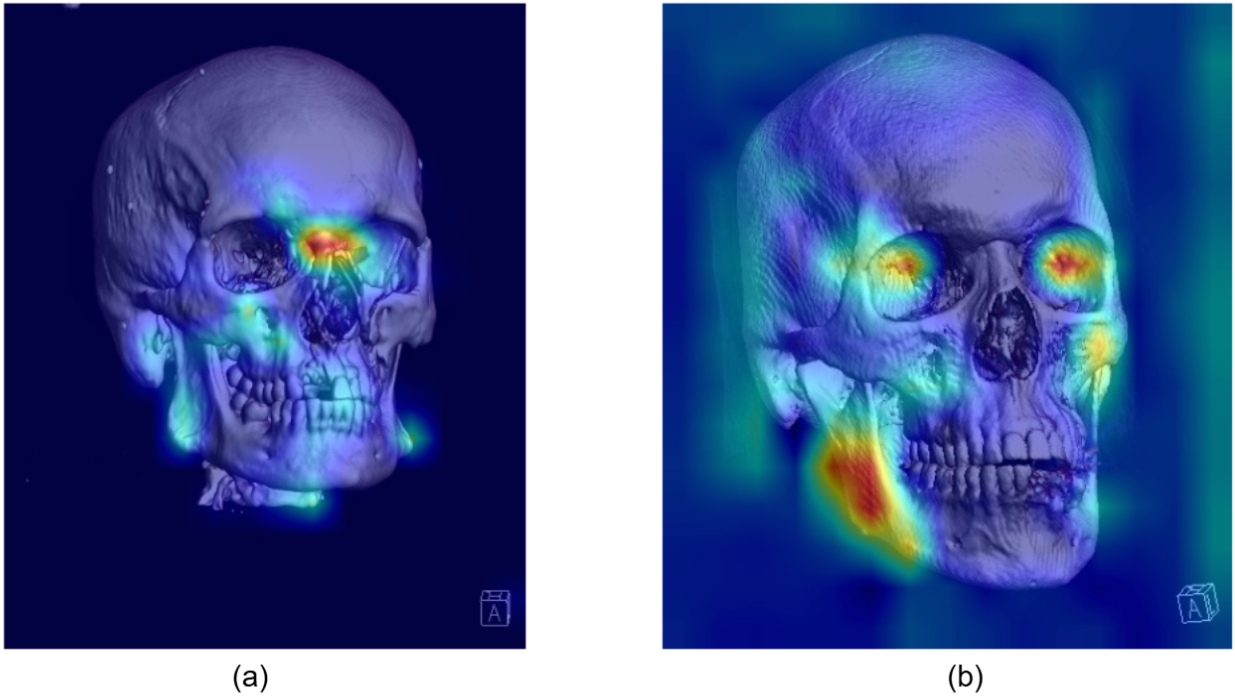}
\caption{\textbf{Male -- (a) Frontonasal angle. (b) Supraorbital margin, mandibular ramus, and angle.}}
\label{fig10}
\end{figure}

The resulting maps highlight the anatomical regions most relevant for classification, following a temperature gradient, where red and yellow areas indicate the highest peak of network attention (the most critical features for the model's decision), transitioning through green, while dark blue areas represent regions with minimal relevance. Comparing these maps with established anthropological knowledge allowed us to verify whether the network identified the same anatomical landmarks used by forensic anthropologists in sexual differentiation (see the Discussion section and examples in Figs.~\ref{fig7} to Fig.~\ref{fig10}).

\subsection{Metric Generation from Validation Batches}

\begin{figure}[!ht]
\centering
\includegraphics[width=1\columnwidth]{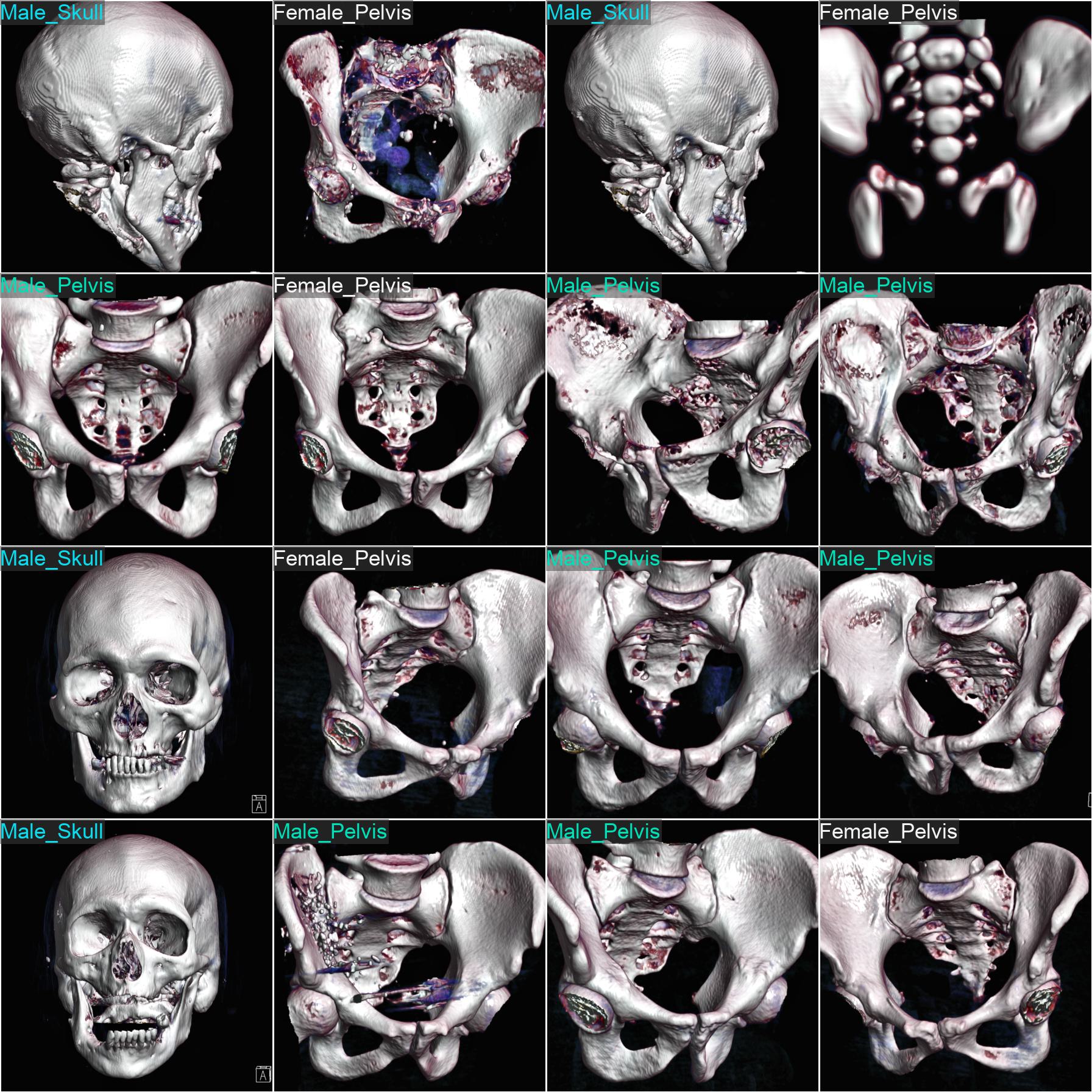}
\caption{\textbf{Label batch representing the correct pre-defined classifications (ground truth).}}
\label{fig:batch_labels}
\end{figure}

\begin{figure}[!ht]
\centering
\includegraphics[width=1\columnwidth]{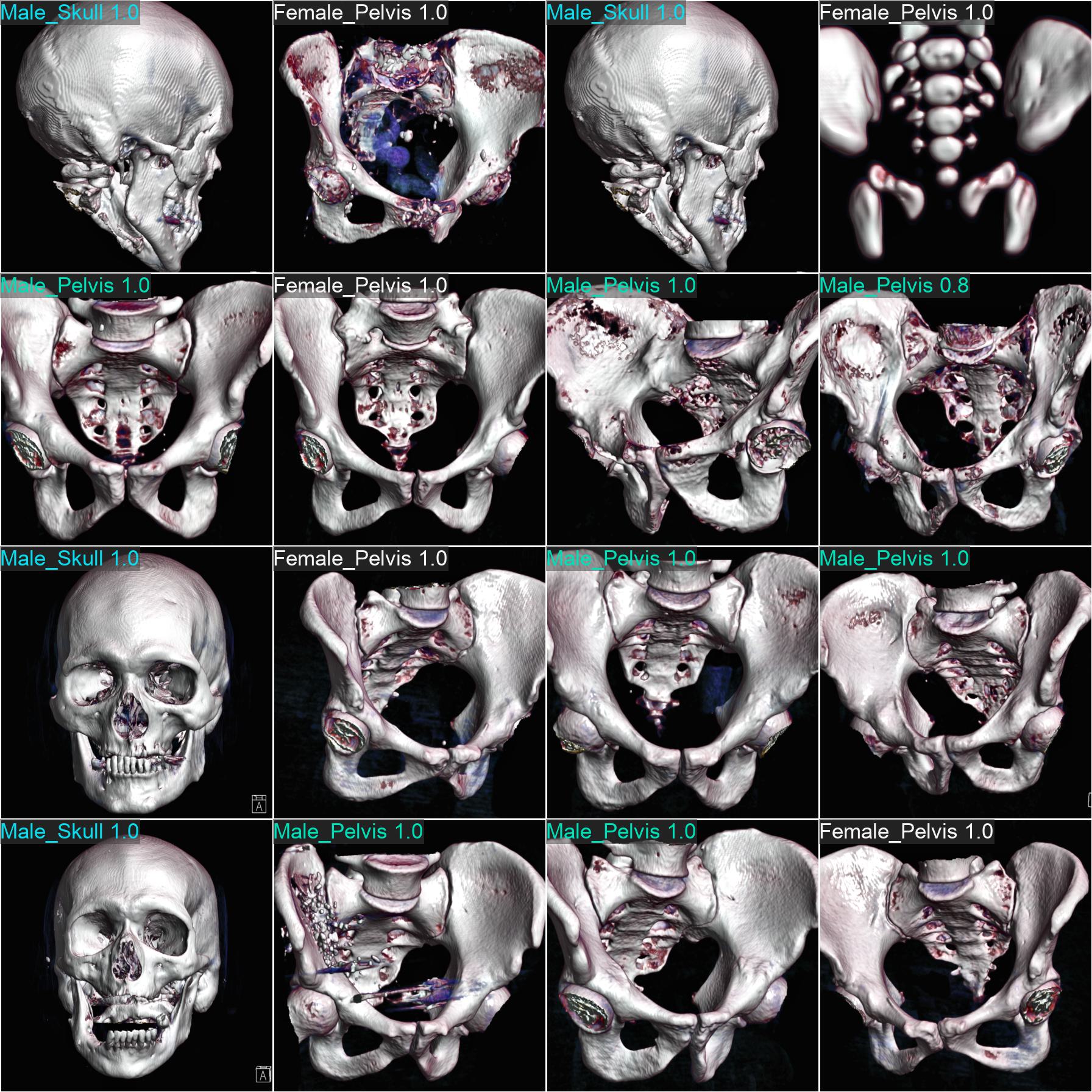}
\caption{\textbf{Prediction batch showing the results of the model's predictions on the validation dataset.}}
\label{fig:batch_predict}
\end{figure}

After completing the training phase, the model's performance is evaluated on the validation subsets that were not seen by the network during that training iteration. To illustrate this validation process, Fig.~\ref{fig:batch_labels} and Fig.~\ref{fig:batch_predict} are presented. They represent a single batch of 16 images and serve as a visual example of how the metrics are generated.

Initially, Fig.~\ref{fig:batch_labels} displays the true labels of each image. This is the correct pre-defined classification in the dataset, which serves as the ground truth for evaluation. 

Subsequently, Fig.~\ref{fig:batch_predict} displays the results of the model's predictions for the exact same batch. Each image classified by the neural network receives a label containing the predicted class and the associated confidence score.

When analyzing this batch, it is noteworthy that all 16 images were correctly classified by the model. Furthermore, the network exhibited extremely high certainty in its predictions, with almost all samples achieving a maximum confidence score of 1.0, and only one sample (second row, fourth column) showing a slightly lower, yet still robust, score of 0.8.


\section{DISCUSSION}



A critical aspect of applying deep learning to forensic anthropology is ensuring that models do not operate as a "black box". The heatmaps presented in Figs.~\ref{fig7} to Fig.~\ref{fig10} provide a visual interpretation of the model's decision-making process, increasing confidence in the reliability of the obtained results. This analysis revealed that the best-performing architectures learned to focus on anatomical regions that strongly correspond to established sexually dimorphic landmarks in forensic anthropology, demonstrating that their predictions are consistent with traditional medico-legal knowledge.

\subsection{Interpretability and Morphological Feature Extraction}

Heatmap analysis revealed that, in the pelvis, the model focused especially on the subpubic angle and the pubic symphysis, correctly identifying increased width and obtuseness in females versus a narrower and more acute configuration in males~\cite{phenice1969}. This finding is consistent with the osteological literature, which recognizes the pelvis, particularly the pubic region, as one of the most dimorphically reliable structures for sex estimation~\cite{klales2012}. The importance of these anatomical landmarks is also reflected in the quantitative results presented in Table ~\ref{tab:model_performances}, where pelvic reconstructions achieved high classification metrics. Such a pattern demonstrates direct scientific convergence with state-of-the-art deep learning studies that likewise identified the pubis (ventral and dorsal surfaces) as a decisive landmark in sex classification~\cite{18deantes}.

Additionally, in cases with partial visualization of the lower limb, the algorithm highlighted the femoral head as a region of interest, as seen in Figure \ref{fig7}. Although the femur remained in some reconstructions without deliberate intent, the model spontaneously identified another dimorphic landmark widely used in forensic anthropology, the diameter of the femoral head~\cite{Femur_head}.

In the skull, the literature describes a set of traits as primary indicators of sexual dimorphism, notably the frontal bone (glabella and supraorbital ridge) and the mandible (mental protuberance and mandibular angle)~\cite{seo, 20deantes}. In female skulls, the heatmaps in Figure \ref{fig9} largely matched these anatomical descriptions, highlighting more gracile sub-regions, such as a smooth glabella, thin orbital margins, and a pointed chin.

A similar pattern was observed in male samples in Figure \ref{fig10}, in which the networks focused on the frontal bones and mandible, emphasizing a thick, projected supraorbital ridge and a quadrangular contour of the mandibular angle/chin. This finding aligns with other deep learning studies in forensic anthropology, which also identified the supraorbital ridge as one of the most discriminative regions for male identification~\cite{seo}.

Additionally, in the heatmaps of Figures \ref{fig9} and \ref{fig10}, the system directed attention to the posterior region of the orbital cavity. Although this area is not among the classic markers of highest discriminative value in traditional forensic anthropology, recent CT-based orbital morphometry studies have described subtle dimorphisms in this region, differences that are difficult for human perception to detect but are detectable by computer vision algorithms~\cite{23deantes}.

Due to the frontal and fronto-lateral image acquisition protocol adopted in this study, certain classical indicators of sexual dimorphism, such as the occipital protuberance located in the posterior region of the skull~\cite{21deantes}, were intentionally excluded from the analysis. Nevertheless, the model successfully differentiated male and female skulls using anatomical features visible within the selected views.

\subsection{Methodological Considerations and Variability of Results}

Different architecture sizes, data augmentation strategies, and class configurations were systematically compared, as shown in Tables~\ref{tab:topsis_results} and~\ref{tab:vikor_results}, using the TOPSIS and VIKOR multi-criteria decision-making methods, respectively. The superiority of the computationally efficient Nano architecture suggests that sexual dimorphism patterns in CT reconstructions are well-defined. Paradoxically, increasing model capacity (Small and Medium architectures) did not improve performance in quaternary classification, indicating a threshold beyond which increased complexity leads to overfitting due to the limited sample size and the highly specific nature of medical scan features.

A critical methodological advancement in this study was the transition from isolated image evaluation to patient-level aggregation. By combining the predictions from 11 rotational projections per anatomical structure via soft voting, diagnostic accuracy improved. This approach effectively neutralizes occasional misclassifications caused by ambiguous angles or localized imaging artifacts, mathematically simulating the holistic, multi-angle inspection performed by human forensic anthropologists.

Furthermore, the superiority of quaternary classification (region + sex) over binary classification (sex only) as shown in Table~\ref{tab:precision_extremes}, reinforces the idea that multi-class decomposition acts as a form of guided feature learning. By separating the skull and pelvis into distinct categories, the network avoids the high intra-class variance of a generalized ''male'' or ''female'' biological class, thereby optimizing the recognition of distinct morphological dimorphisms. In binary classification, however, the Medium architecture exhibited superior performance, as expected, since combining different bones into the same category requires greater representation capacity for adequate generalization.

Regarding data augmentation strategies, the customized approach, contrary to expectations, reduced performance in most scenarios. Possibly, the applied transformations, although designed to preserve bone morphology, inadvertently compromised subtle anatomical features crucial for sexual differentiation. In contrast, the standard YOLO approach, optimized on massive, diverse datasets, achieved better empirical performance in our experiments.

\subsection{Practical Implications}

The high patient-level diagnostic accuracy, alongside the robust classification performance obtained for pelvic structures as shown in Table~\ref{tab:model_performances}, aligns with the upper bound of the literature, surpassing studies using traditional machine learning~\cite{22deantes} and analog CT-based sex differentiation approaches~\cite{21deantes}. This performance stems from deep learning's ability to analyze the image as a whole and extract global discriminative patterns without being restricted to predefined measurements or parameters.

The findings indicate that deep learning can serve as an objective, rapid, and reproducible support tool in routine forensic practice, particularly in cases involving anatomical disfigurements or trauma-related imaging artifacts, where conventional morphological assessment may become more challenging or less reliable.

Regarding practical applicability, the system should act as a supporting instrument, not a substitute, for the forensic expert. Predictions are probabilistic and require individualized validation. Furthermore, given that automated decisions do not allow for direct attribution of responsibility, algorithmic outputs should not, in isolation, form the basis for judicial action. This limitation necessitates mechanisms for traceability and auditability (versioning, logs), explainability (such as heatmaps), and report standardization.

In summary, the observed performance, the feasibility of sample expansion, and the concrete prospect of progressive refinement support the evolution toward forensic support software with governance compatible with the digital chain of custody, capable of streamlining triage in high-demand or taphonomic-degradation contexts, while preserving the expert's decision-making role.

\section{Conclusion}

This study demonstrated the high efficacy of deep learning architectures, enhanced by transfer learning, for automated biological sex determination using 2D projections extracted from post-mortem pelvic and cranial CT scans. By employing an optimized quaternary classification strategy and a patient-level aggregation scheme, the proposed methodology achieved an exceptional overall diagnostic accuracy of 95.65\%. Furthermore, the model achieved macro-averaged precision of 97.22\%, recall of 92.86\%, and F1-score of 94.36\%, correctly classifying all pelvic structures in the test set.

These results directly address the primary objectives of this work, delivering its main contributions by validating deep learning architectures on a diverse dataset of real-world forensic cases featuring significant anatomical disfigurements and trauma-related artifacts, conducting a comprehensive comparative analysis of the discriminative power between pelvic and cranial 2D projections, and providing a robust assessment of the generalization capabilities of transfer learning in the presence of forensic-specific image noise. 

Ultimately, this research establishes that deep learning can serve as a highly reliable, objective, and high-speed assistive tool in routine forensic practice. While future studies are encouraged to expand the dataset, the current framework demonstrates technical feasibility, paving the way to streamline forensic triage and support medical examiners.

\section{Acknowledgment}
The authors thank Dr. Rafaella Marques Barbosa, Manager of the Aristoclides Teixeira Institute of Forensic Medicine (IMLAT), for enabling institutional access and for providing crucial support in acquiring the forensic tomographic dataset.




\bibliographystyle{IEEEtran}
\bibliography{references}

\begin{IEEEbiography}[{\includegraphics[width=1in,height=1.25in,clip,keepaspectratio]{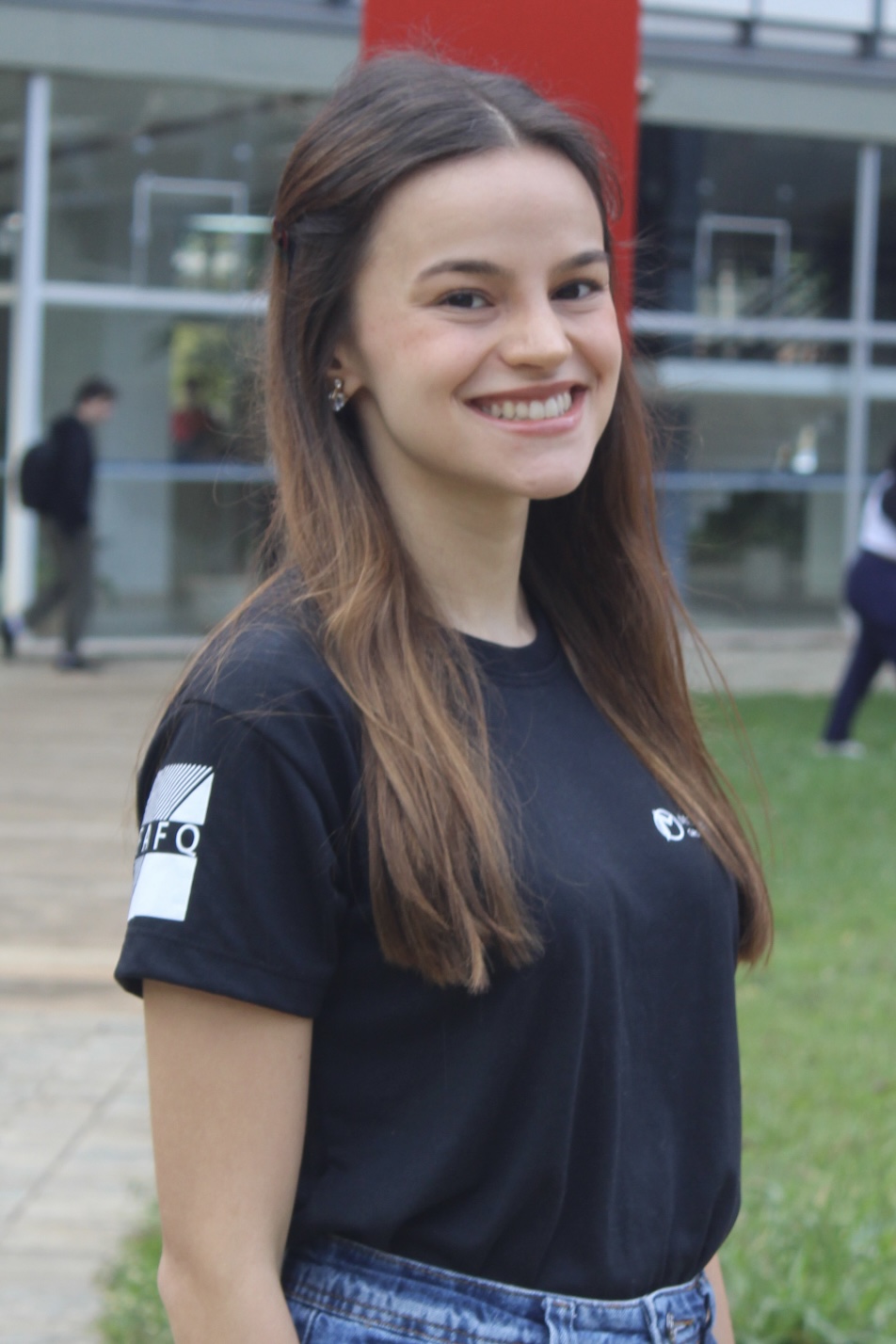}}]{Giovanna Herculano Tormena}
is currently working toward the B.S. degree in Mechatronics Engineering. She has experience in the fields of computer vision and artificial intelligence, with previous research work published in these areas. Her research interests include computer vision, machine learning, and their applications in mechatronic systems and robotics.
\end{IEEEbiography}

\begin{IEEEbiography}[{\includegraphics[width=1in,height=1.25in,clip,keepaspectratio]{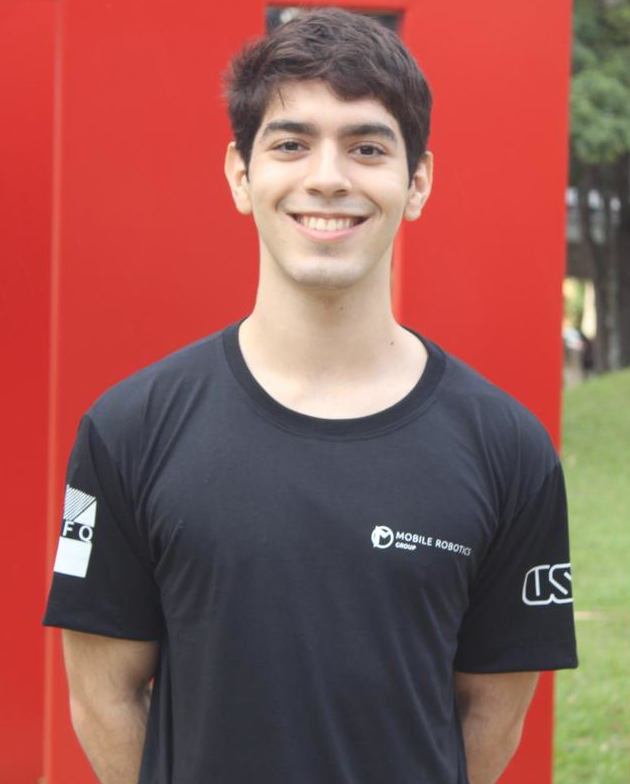}}]{Davi Nascimento Araújo}
is currently pursuing a B.Sc. degree in Computer Science at the University of São Paulo (USP), acting as a undergraduate researcher in the Mobile Robotics Group with interests centered on computer vision, machine learning, and robotics.
\end{IEEEbiography}

\begin{IEEEbiography}[{\includegraphics[width=1in,height=1.25in,clip,keepaspectratio]{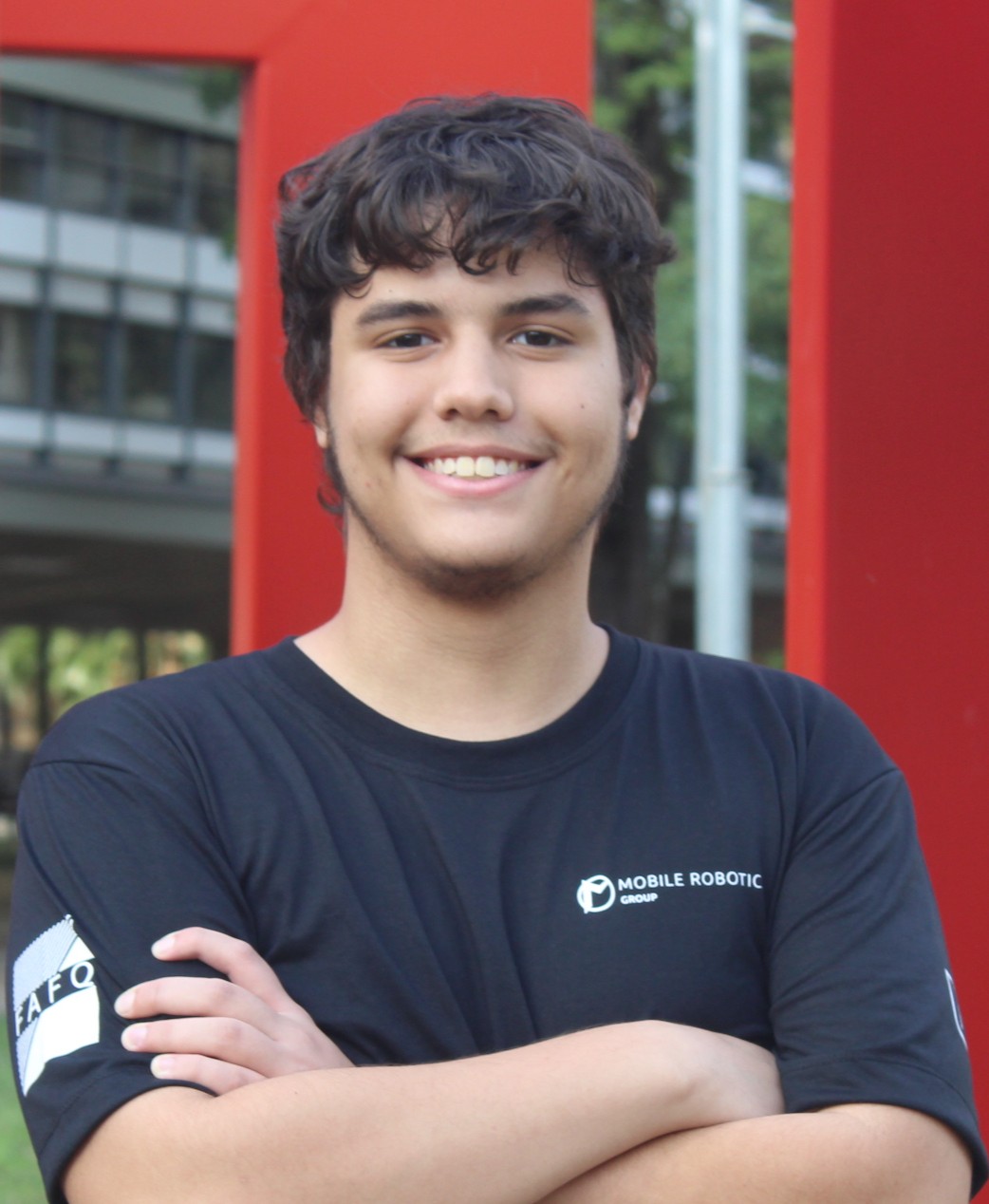}}]{Germano Coimbra Soares de Carvalho}
is currently working toward the B.S. degree in Mechatronics Engineering. He has experience in the fields of computer vision and artificial intelligence. His research interests include computer vision, machine learning, and their applications in mechatronic systems and robotics.
\end{IEEEbiography}

\begin{IEEEbiography}[{\includegraphics[width=1in,height=1.25in,clip,keepaspectratio]{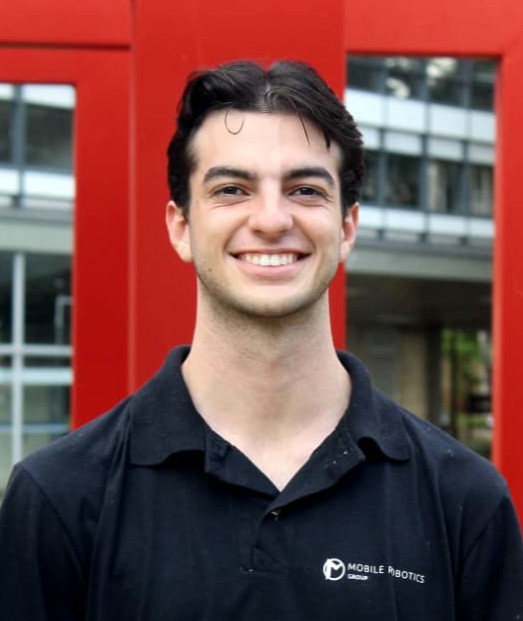}}]{Gustavo Bruno Centenaro}
is currently pursuing the Bachelor of Engineering degree in electrical engineering with an emphasis in electronics with the University of São Paulo (USP). He is an Undergraduate Researcher with the USP's Mobile Robotics Group, where he contributes to advancements in the field of mobile robotics.
\end{IEEEbiography}

\begin{IEEEbiography}[{\includegraphics[width=1in,height=1.25in,clip,keepaspectratio]{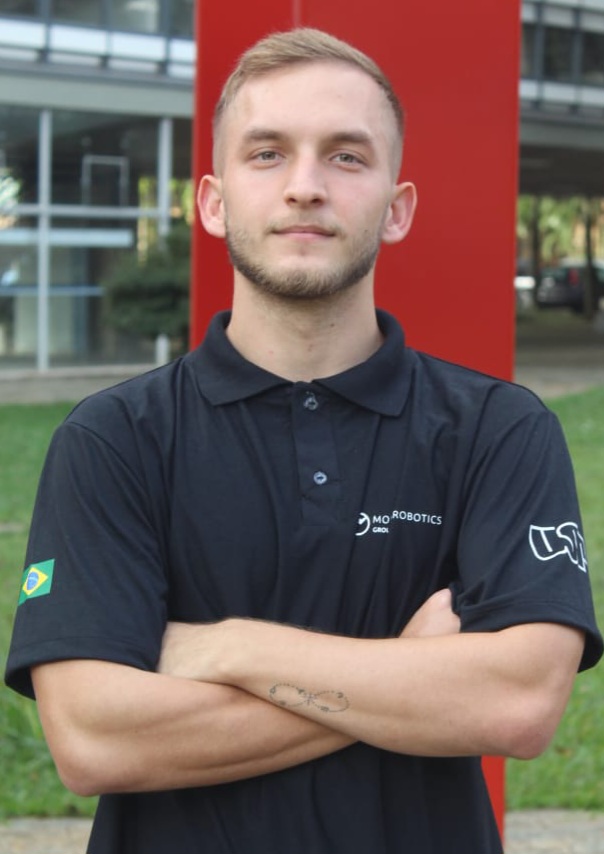}}]{Rafael Janowski Pozzer}
is currently pursuing a B.S. degree in Mechatronics Engineering at the University of São Paulo (USP), acting as an undergraduate researcher in the Mobile Robotics Group. His research interests include computer vision and machine learning.
\end{IEEEbiography}

\begin{IEEEbiography}[{\includegraphics[width=1in,height=1.25in,clip,keepaspectratio]{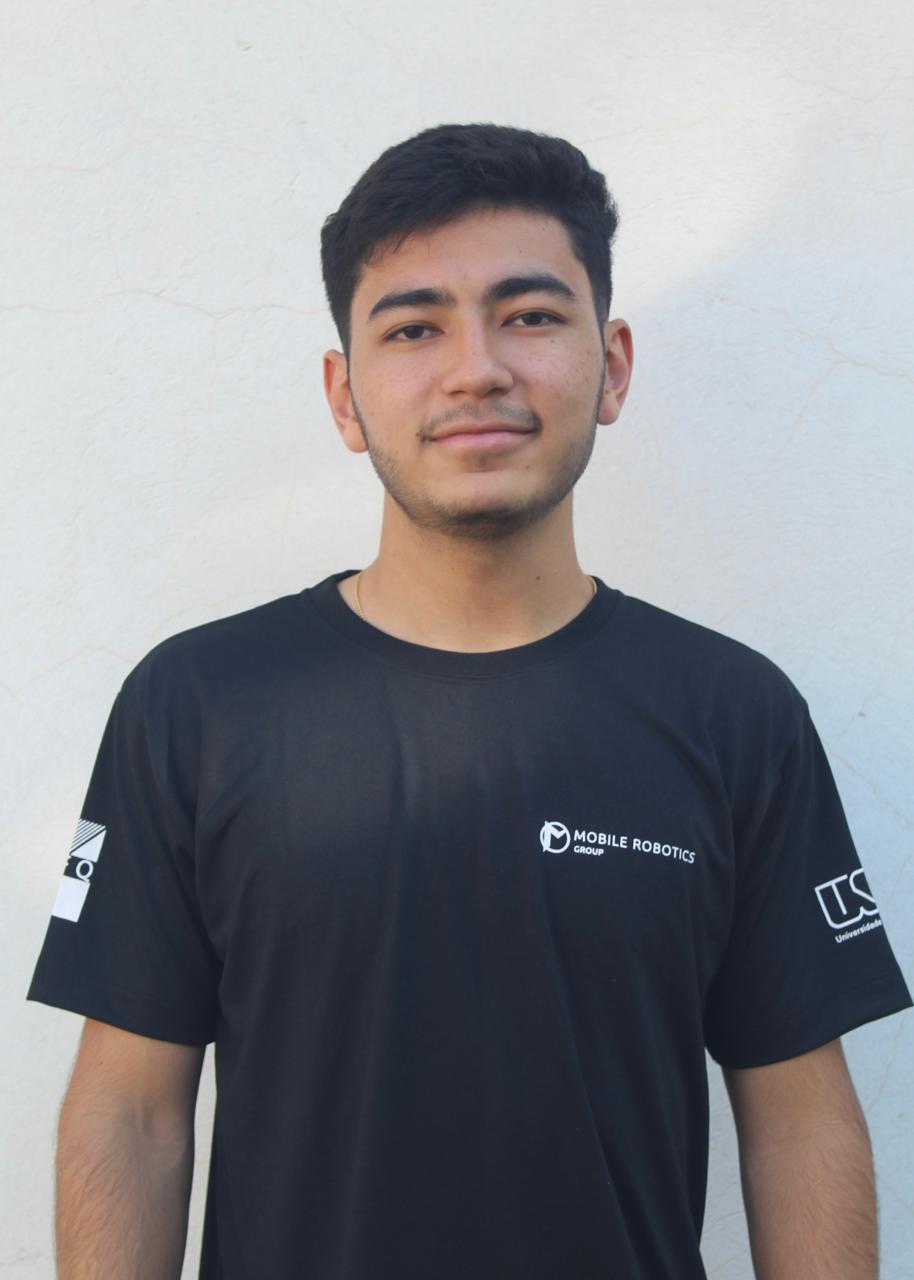}}]{Rodrigo Akira Azevedo Kurosawa}
is currently pursuing the Bachelor of Engineering degree in Computer Engineering at the University of São Paulo (USP). He is an Undergraduate Researcher working in the fields of robotics, machine learning, and computer vision, contributing to the development of intelligent and autonomous systems.
\end{IEEEbiography}

\begin{IEEEbiography}[{\includegraphics[width=1in,height=1.25in,clip,keepaspectratio]{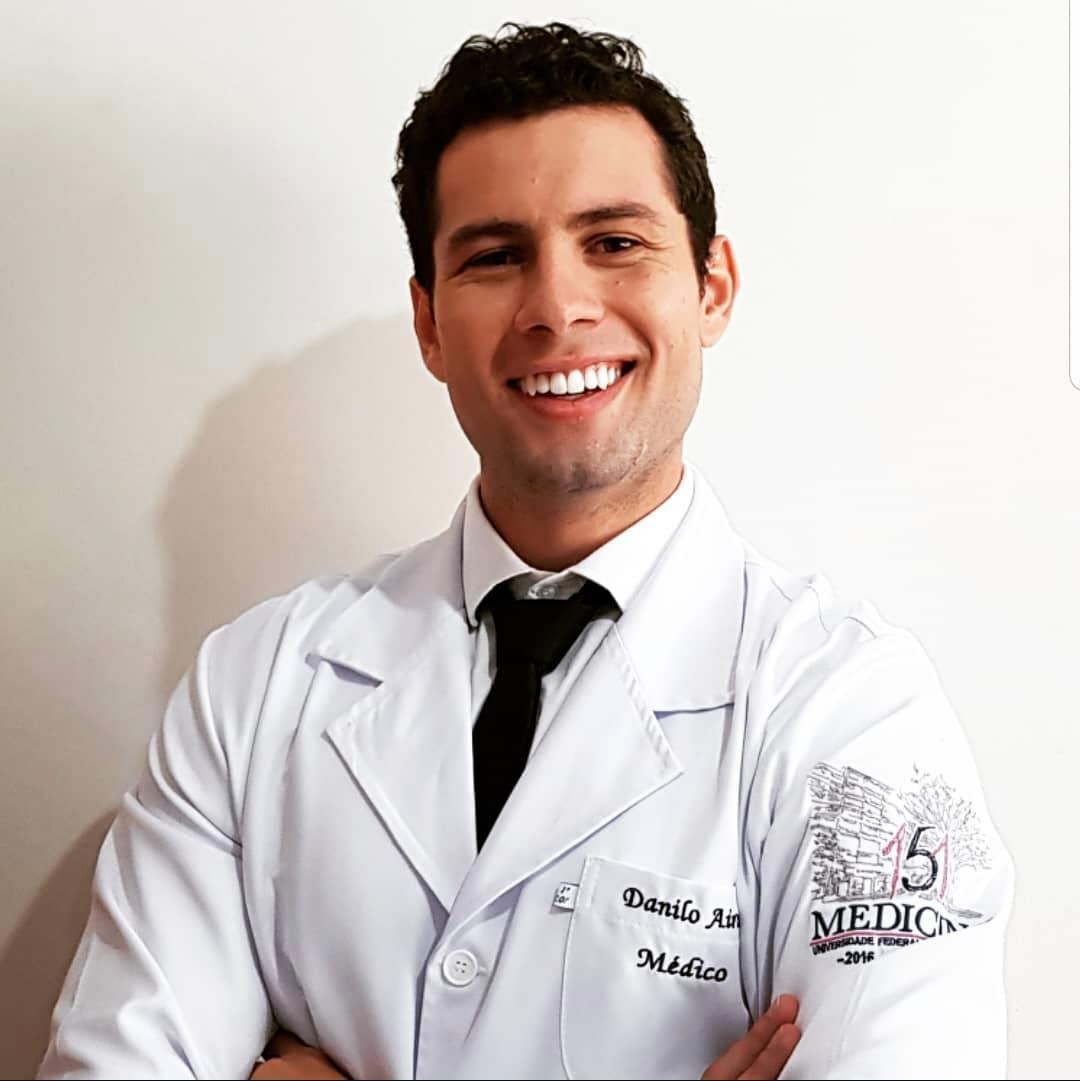}}]{Danilo Aires Alves}
is a physician currently working as a medical examiner (coroner), with a background as a former medical officer in the Brazilian Army. He holds a postgraduate degree in legal medicine and has experience in clinical practice and forensic medicine. His professional interests include legal medicine, forensic sciences, and the application of artificial intelligence in forensic analysis.
\end{IEEEbiography}

\begin{IEEEbiography}[{\includegraphics[width=1in,height=1.25in,clip,keepaspectratio]{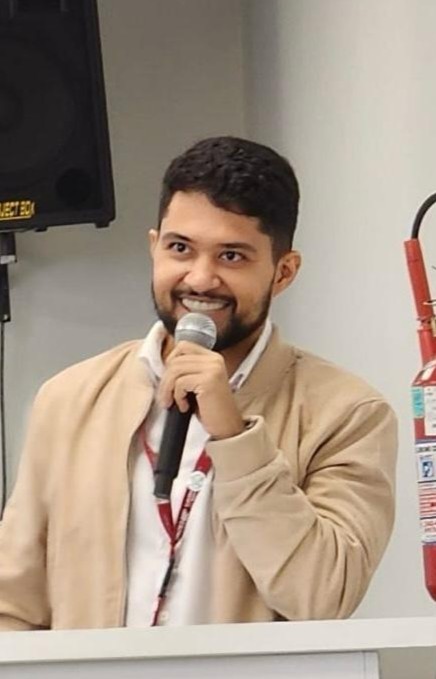}}]{Filipe Thiago Xavier de Campos}
is a neuroradiologist with a Master’s degree in Health Sciences from Universidade Federal de Goiás. He is a full member of the Sociedade Brasileira de Neurorradiologia and the Colégio Brasileiro de Radiologia. He earned his medical degree from the Universidade Federal de Goiás.
Dr. Campos currently serves as Head of the Medico-Legal Specialty Department at the Instituto Médico Legal de Goiânia and Head of the Imaging Department at the Hospital das Clínicas da UFG. His main areas of interest include forensic radiology, neuroradiology, head and neck imaging, and artificial intelligence applied to healthcare, particularly in radiology.
\end{IEEEbiography}

\begin{IEEEbiography}[{\includegraphics[width=1in,height=1.25in,clip,keepaspectratio]{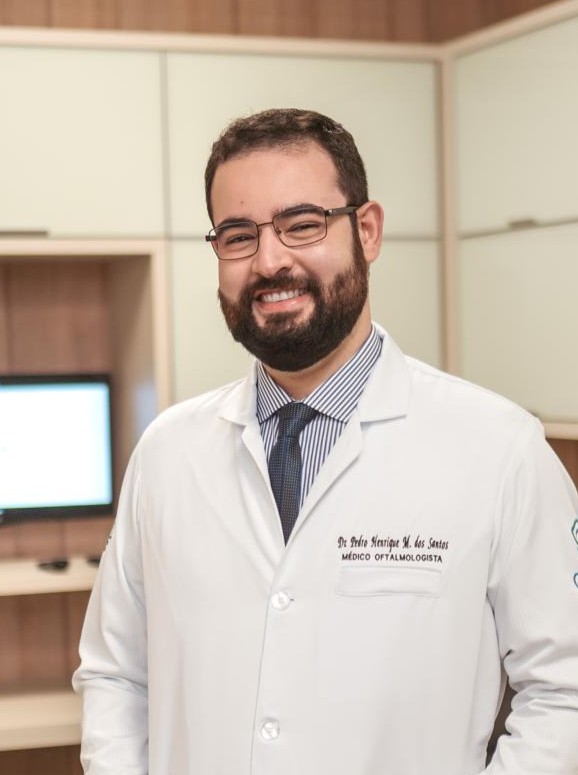}}]{Pedro Henrique Macedo dos Santos}
is a medical ophthalmologist, serving as staff physician and coordinator of the Ophthalmology Department at Hospital e Maternidade Dona Íris, in Goiânia, Goiás, Brazil. He received his Medical Degree from the Federal University of Triângulo Mineiro and completed his Ophthalmology residency and Fellowship in Cornea and External Ocular Diseases at the Federal University of Goiás. He also serves as a Forensic Medical Examiner at the Scientific Police of Goiás and as a Court-Appointed Expert Witness for the Court of Justice of the State of Goiás. His research interests include forensic medicine, ophthalmological expert evaluation, cornea and external ocular diseases, and their applications in the medico-legal context.
\end{IEEEbiography}

\begin{IEEEbiography}[{\includegraphics[width=1in,height=1.25in,clip,keepaspectratio]{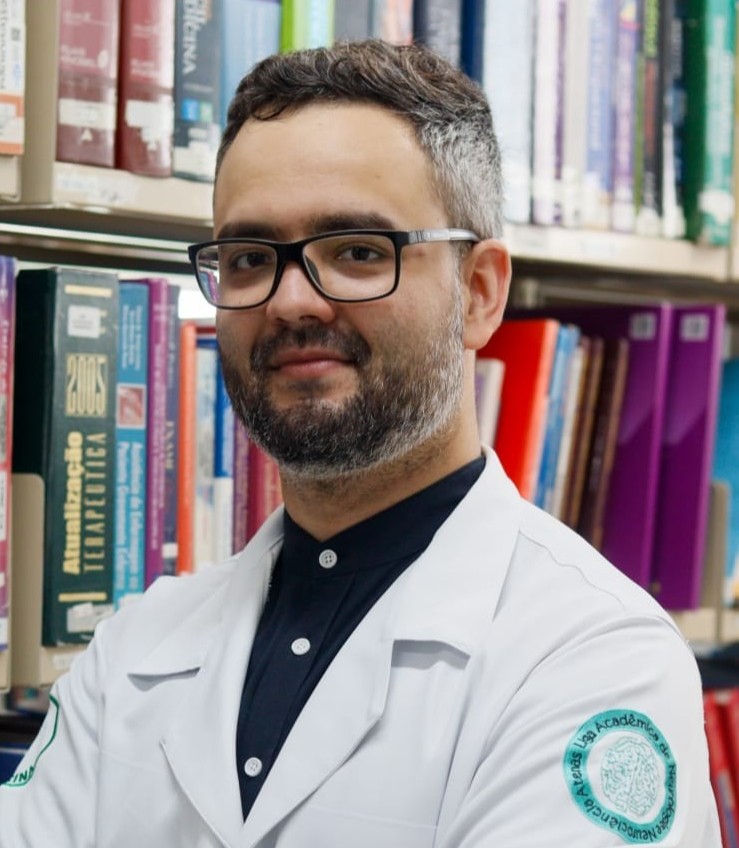}}]{Pedro Augusto Prado Mota}
is currently finishing his medical radiology training and already works as a medical coroner.
He has experience in emergency medicine and medical imaging. He has interests in legal medicine, medical radiology, forensic radiology, and artificial intelligence.
\end{IEEEbiography}

\begin{IEEEbiography}[{\includegraphics[width=1in,height=1.25in,clip,keepaspectratio]{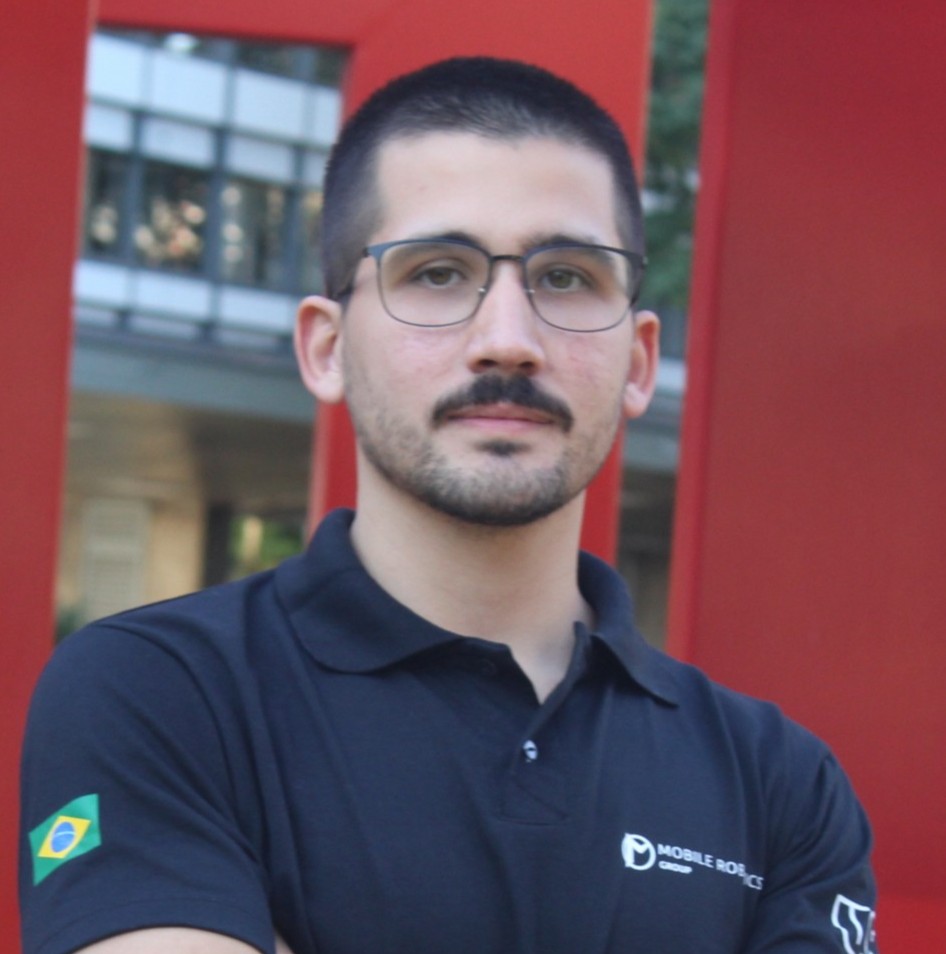}}]{Ricardo V. Godoy} received the Bachelor of Engineering in Mechatronics Engineering in 2019, followed by M.Sc. in 2021 in Mechanical Engineering from the University of São Paulo, São Carlos, Brazil. He received his PhD in Mechatronics Engineering with the New Dexterity Research Group of the University of Auckland, New Zealand, where he worked on the analysis and development of novel Human-Machine Interfaces (HMI) for the control of robotic and bionic devices while focusing on the challenges and limitations in the use of HMI for robust grasping and decoding of dexterous, in-hand manipulation tasks. He is currently pursuing his postdoc at the University of São Paulo, São Carlos, Brazil, working towards the development of robotic frameworks for inspection and automation, focusing on manipulation and loco-manipulation frameworks.
\end{IEEEbiography}

\begin{IEEEbiography}[{\includegraphics[width=1in,height=1.25in,clip,keepaspectratio]{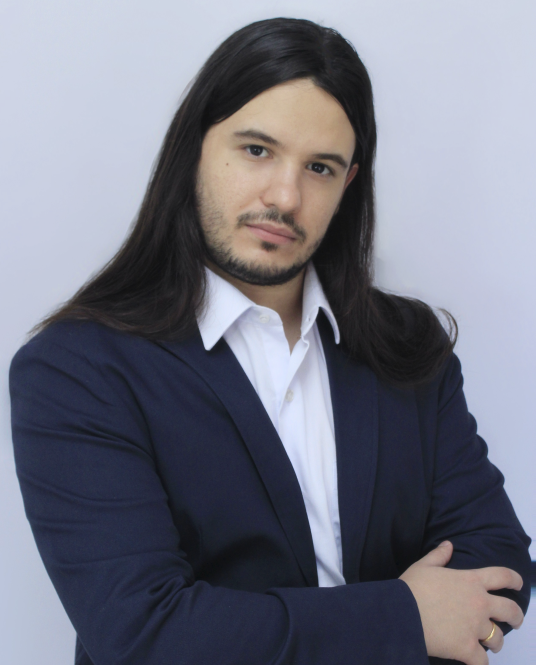}}]{João Manoel Herrera Pinheiro}
received the M.Sc. degree in Mechanical Engineering, with a focus on computer vision and robotics, and the B.Sc. degree in Mechatronics Engineering from the University of São Paulo (USP). He is currently pursuing the D.Sc. degree in Electrical Engineering, with research interests centered on computer vision, machine learning, and image and signal processing. He has also completed two postgraduate specialization programs: Didactic–Pedagogical Processes for Distance Learning (UNIVESP) and Software Engineering (USP). He serves as a reviewer for several international journals, including IEEE Latin America Transactions, Scientific Data (Nature), Artificial Intelligence (IBERAMIA), and the Journal of the Brazilian Computer Society.
\end{IEEEbiography}

\begin{IEEEbiography}[{\includegraphics[width=1in,height=1.25in,clip,keepaspectratio]{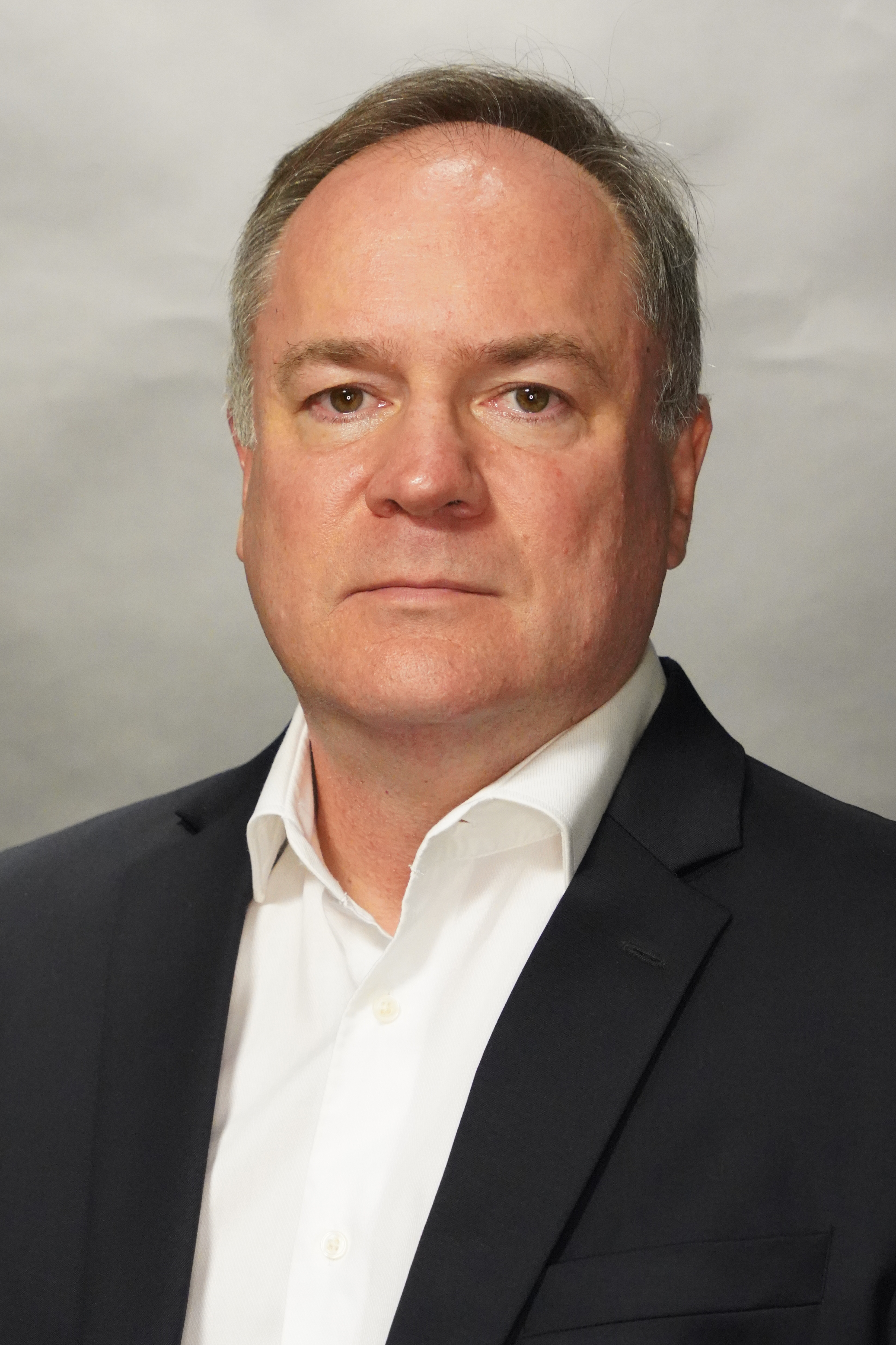}}]{Marcelo Becker} received the B.Sc. degree in Mechanical Engineering (Mechatronics) from the University of São Paulo, Brazil, in 1993, and the M.Sc. and D.Sc. degrees in Mechanical Engineering from the University of Campinas, Brazil, in 1997 and 2000, respectively. He was a visiting researcher at ETH Zürich and did a sabbatical at EPF Lausanne, Switzerland. He is currently an Associate Professor at the University of São Paulo and coordinates the Mobile Robotics Group and the USP Center of Robotics (CRob). His research interests include mobile robotics, automation, perception systems, and mechatronic design for applications in agriculture and industrial automation.
\end{IEEEbiography}






\EOD

\end{document}